%% file: 0_all.tex
\documentclass[acmsmall,screen]{acmart}

\usepackage{multirow}
\usepackage{amsmath}
\usepackage{amsthm}
\usepackage{todonotes}
\usepackage{threeparttable}
\usepackage{chapterbib}
\makeatletter
\newcommand{\biggg}{\bBigg@{1.2}}  
\def\bigggl{\mathopen\biggg}

\makeatother
\AtBeginDocument{%
  \providecommand\BibTeX{{%
    \normalfont B\kern-0.5em{\scshape i\kern-0.25em b}\kern-0.8em\TeX}}}

\setcopyright{acmcopyright}
\copyrightyear{2023}
\acmYear{2023}
\acmDOI{XXXXXXX.XXXXXXX}

\begin{document}

\title{State of the Art and Potentialities of Graph-level Learning}

\author{Zhenyu Yang, Ge Zhang, Jia Wu, Jian Yang, Quan Z. Sheng, Shan Xue}
\affiliation{%
  \institution{Macquarie University}
  \city{Sydney}
  \country{Australia}}
\email{zhenyu.yang3@hdr.mq.edu.au,ge.zhang5@hdr.mq.edu.au, jia.wu@mq.edu.au, jian.yang@mq.edu.au,
michael.sheng@mq.edu.au,
emma.xue@mq.edu.au
}

\author{Chuan Zhou}
\affiliation{%
  \institution{Chinese Academy of Sciences}
  \city{Beijing}
  \country{China}}
\email{zhouchuan@amss.ac.cn}

\author{Charu Aggarwal}
\affiliation{%
  \institution{IBM T. J. Watson Research Center}
  \city{New York}
  \country{USA}}
\email{charu@us.ibm.com}

\author{Hao Peng}
\affiliation{%
  \institution{Beihang University}
  \city{Beijing}
  \country{China}}
\email{penghao@buaa.edu.cn}

\author{Wenbin Hu}
\affiliation{%
  \institution{Wuhan University}
  \city{Wuhan}
  \country{China}}
\email{hwb@whu.edu.cn}

\author{Edwin Hancock}
\affiliation{%
  \institution{University of York}
  \city{York}
  \country{United Kingdom}}
\email{edwin.hancock@york.ac.uk}

\author{Pietro Liò}
\affiliation{%
  \institution{University of Cambridge}
  \city{Cambridge}
  \country{United Kingdom}}
\email{Pietro.Lio@cl.cam.ac.uk}

\renewcommand{\shortauthors}{Zhenyu Yang et al.}
\begin{abstract}
Graphs have a superior ability to represent relational data, like chemical compounds, proteins, and social networks. Hence, graph-level learning, which takes a set of graphs as input, has been applied to many tasks including comparison, regression, classification, and more. Traditional approaches to learning a set of graphs heavily rely on hand-crafted features, such as substructures. But while these methods benefit from good interpretability, they often suffer from computational bottlenecks as they cannot skirt the graph isomorphism problem. Conversely, deep learning has helped graph-level learning adapt to the growing scale of graphs by extracting features automatically and encoding graphs into low-dimensional representations. As a result, these deep graph learning methods have been responsible for many successes. Yet, there is no comprehensive survey that reviews graph-level learning starting with traditional learning and moving through to the deep learning approaches. This article fills this gap and frames the representative algorithms into a systematic taxonomy covering traditional learning, graph-level deep neural networks, graph-level graph neural networks, and graph pooling. To ensure a thoroughly comprehensive survey, the evolutions, interactions, and communications between methods from four different branches of development are also examined. This is followed by a brief review of the benchmark data sets, evaluation metrics, and common downstream applications. 
The survey concludes with a broad overview of 12 current and future directions in this booming field.
\end{abstract}



\received{20 February 2007}
\received[revised]{12 March 2009}
\received[accepted]{5 June 2009}

\maketitle

\include{0-main}

\clearpage
\appendix
\include{12-appendix}

\end{document}

%% file: 0-main.tex
\input{1-Introduction}

\input{2-Definitions}
\input{3-Development_and_Taxonomy}

\input{4-Traditional_Learning_Techniques}

\input{5-GLNN-DNN}

\input{6-GLNN-GNN}

\input{7-GLNN-Pooling}

\input{8-Datasets_and_Evaluations}
\input{9-Applications}

\input{10-Future_Directions}

\input{11-Conclusions}

\bibliographystyle{ieeetr}
\bibliography{acm}

%% file: 1-Introduction.tex
\section{Introduction}

Research into graph-structured data started with the Konigsberg bridge problem \cite{euler1741solutio} in the 18 century, that is: ``\textit{How can we design a path among seven bridges in Konigsberg city that crosses each bridge only once}?'' Through modeling seven bridges into a graph in which nodes represent the junctions between bridges and edges represent bridges, the Konigsberg bridge problem is proved unsolvable. 
Since then, graph-structured data has become an indispensable tool for exploring the world.
In reality, researchers can model millions of molecules in which each presents a graph to analyze molecular properties \cite{wu2018moleculenet}.
Such case learning the underlying semantics among a set of graphs is graph-level learning.

Mining the underlying rules among a set of graphs is tough hard as graphs are irregular with an unfixed number of disordered nodes and varied structural layouts.
A long-standing challenge in graph-level learning, the graph isomorphism problem, is ``\textit{How to determine whether two graphs are completely equivalent or isomorphic\footnote{Two graphs $\mathcal G_1$ and  $\mathcal G_2$ are isomorphic if the following two conditions are met: (1) There exists matching between nodes in $\mathcal G_1$ and $\mathcal G_2$; (2) Two nodes are linked by an edge in $\mathcal G_1$ \textit{iff} the corresponding nodes are linked by an edge in $\mathcal G_2$.}}?'' 
An enormous number of studies \cite{harary1969four, weisfeiler1968reduction,mckay1981practical} focused on this question and concerned it as a candidate for NP-immediate until a quasi-polynomial-time solution was proposed in 2016 \cite{babai2016graph}. 
To tackle the struggle in this area, tremendous efforts have been made involving traditional methods and deep learning.

Generally, traditional graph-level learning builds the architecture upon handcrafted features (e.g., random walk sequences \cite{gartner2003graph}, 
frequently occurring substructure \cite{inokuchi2000apriori}) and classical machine learning techniques (e.g.,  support vector
machine).
This paradigm is human-interpretable but is usually restricted to simple small graphs rather than reality large networks.
This is because traditional methods cannot bypass the graph isomorphism problem, the predefined features require to preserve the isomorphism between graphs, i.e., mapping isomorphism graphs to the same features.
On the contrary, deep learning techniques break the shackles by training the network to automatically learn non-linear and low-dimensional features.
This makes deep neural networks bring new benchmarks for state-of-the-art performance and support the ever-increasing size of graph data.
The fly in the ointment is the black-box nature of deep learning, which leads to compromised trustworthiness. An emerging trend is to develop reliable graph-level learning techniques that own the advantages of neural networks and traditional methods. 

Benefiting from these techniques, graph-level learning has applications and promise in many fields.
Wang \textit{et al.} \cite{wang2022molecular} took graphs of molecules, where the nodes denote atoms and the edges represent chemical bonds, and performed graph regression as a way of predicting molecular proprieties to help discover more economical crystals. 
In another study, a graph generation task based on a series of protein graphs was used to produce graphs of proteins with specific functions to support drug discovery \cite{vamathevan2019applications}. 
Likewise, graph classification with brain graphs has the potential to distinguish brain structures with neurological disorders from those of healthy individuals \cite{lanciano2020explainable}.

\begin{figure*}
    \centering
    \includegraphics[width=1\textwidth]{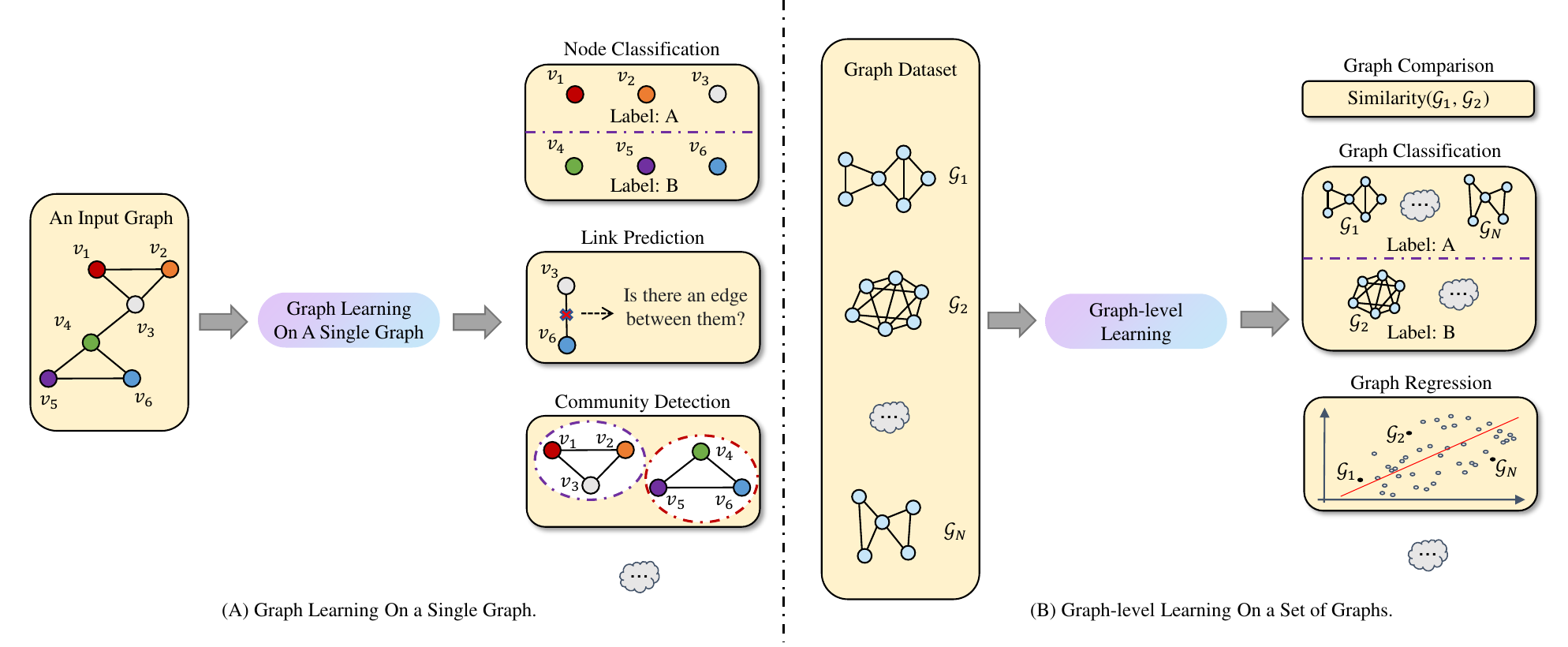}
    \caption{Toy examples of graph learning on a single graph and graph datasets.}
    \label{intro_graphlevel}
\end{figure*}

The success of applications qualifies the huge potential of graph-level learning, which raises a practical demand to comprehensively survey this field spanning both traditional and deep learning within the vast amount of literature.
There are surveys on learning graph-structured data. 
However, these reviews suffer from two main disadvantages.
First, most existing surveys concentrate on articles that explore the node/edge/substructures in a single graph, such as network embedding \cite{cui2018survey}, community detection \cite{su2022comprehensive,10.5555/3491440.3492133}, anomaly detection \cite{ma2021comprehensive}, and graph neural networks \cite{wu2020comprehensive,zhang2020deep}. Graph-level learning is treated as a by-product taking up a subsection or less. The differences between graph learning on a single graph and graph-level learning are illustrated in Fig. \ref{intro_graphlevel}.
Second, graph-level learning is only investigated from a single perspective, such as graph kernels \cite{kriege2020survey} or graph pooling \cite{liu2022graph}. 
As such, the surveys have not covered a broad width and overlook the interactions between different graph-level learning techniques, e.g., adopting traditional techniques to empower GL-GNNs (see sections \ref{subgraph-based GL-GNNs} and \ref{kernel-based GL-GNNs}).

\begin{figure*}
    \centering
   \includegraphics[width=1\textwidth]{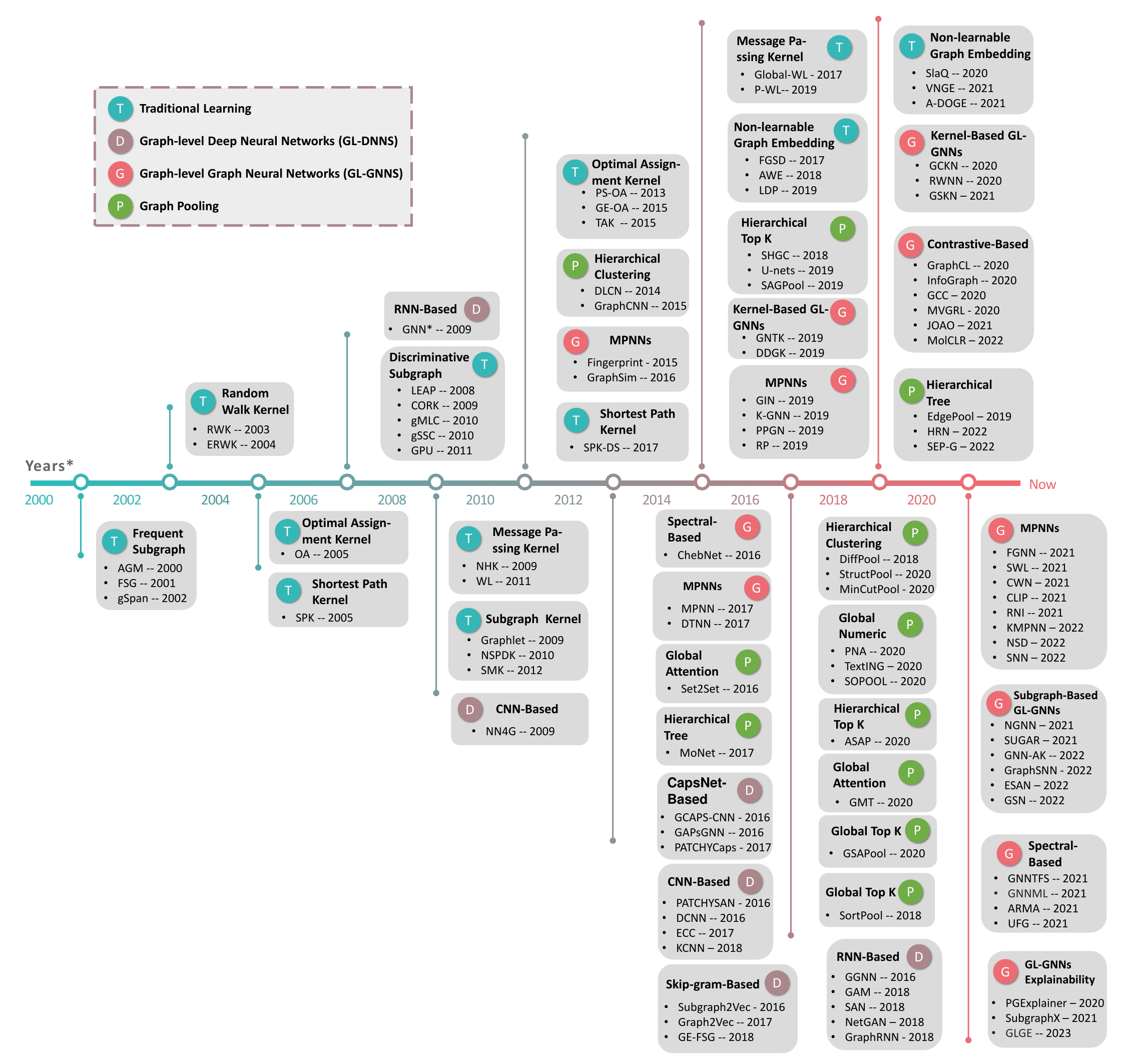}
    \caption{The timeline of graph-level learning in terms of four mainstream techniques.}
    \label{timeline}
\end{figure*}

To the best of our knowledge, this is the first comprehensive survey of graph-level learning that spans both traditional methods and deep learning-based techniques (i.e. GL-DNNs, GL-GNNs, and graph pooling). 
This article exhaustively depicts the mainstream techniques in different periods of graph-level learning (see Fig. \ref{timeline}), and further discusses the evolutions, interactions, and communications between them. 
Thus, the contributions of this survey include:
\begin{itemize}
    \item \textbf{A comprehensive taxonomy:} We propose a comprehensive taxonomy for graph-level learning techniques. Specifically, our taxonomy covers graph-level learning through both traditional and deep learning methods.
    \item \textbf{An in-depth review:} Over four categories, we summarize the representative algorithms, make comparisons, and discuss the contributions and limitations of existing methods. 
    \item \textbf{Abundant resources:} This survey provides readers with abundant resources of graph-level learning, including information on the state-of-the-art algorithms, the benchmark datasets for different domains, fair evaluation metrics for different graph-level learning tasks, and practical downstream applications. The repository of this article is available at \href{https://github.com/ZhenyuYangMQ/Awesome-Graph-Level-Learning}{https://github.com/ZhenyuYangMQ/Awesome-Graph-Level-Learning}.
    \item \textbf{Future directions:} We identify 12 important future directions in the graph-level learning area. 
\end{itemize}

%% file: 2-Definitions.tex
\section{Definitions}


This section, provides some definitions that are essential to understanding this paper. Bold lowercase characters (e.g., $\mathbf x$) are used to denote vectors.
Bold uppercase characters (e.g., $
\mathbf X$) are used to denote matrices. 
Plain uppercase characters (e.g., $\mathcal V$) are used to denote mathematical sets, and lowercase -italic characters (e.g., $n$) are used to denote constants.
\begin{definition}
\textbf{(Graph)}: A graph can be denoted as $\mathcal G = (\mathcal V, \mathcal E)$, where the node set $\mathcal V$ having $n$ nodes (also known as vertices) and the edge set $\mathcal E$ having $m$ edges. In an undirected graph, $\mathcal E_{u,v} = \{u,v\} \in \mathcal E$ represents that there is an edge connecting nodes $u$ and $v$, where $u\in\mathcal V$ and $v\in \mathcal V$. If $\mathcal G$ is unweighted, we use an adjacency matrix $\mathbf A \in \{0,1\}^{n \times n}$ to describe its topological structure, where $\mathbf {A} _{u,v} = 1$ if $\mathcal E_{u,v} \in \mathcal E$, otherwise, 0. If $\mathcal G$ is weighted, the value of $\mathbf {A} _{u,v}$ refers to the weight value of $\mathcal E_{u,v}$. $\mathbf X\in \mathbb R^{n\times f}$ is the node attribute matrix and a node $u\in \mathcal V$ can be described by an attribute vector $\mathbf x_{u} \in \mathbb R^f$. Similarly, the edge feature matrix is denoted as $\mathbf S\in \mathbb R^{m\times d}$, where $\mathbf s_{u, v}\in \mathbb R^{d}$ describes the edge $\mathcal E_{u, v}\in \mathcal E$. Unless otherwise specified, the graphs in this paper are undirected attributed graphs.
\end{definition} 

\begin{definition}
\textbf{(Graph Dataset)}: A graph dataset $\mathbb G$ is composed of $N$ graphs, where $\mathbb G=\{\mathcal G_1, ..., \mathcal G_N\}$.
\end{definition}

\begin{definition}
\textbf{(Subgraph/Substructure)}: A graph $g_m = (\mathcal V_{g_m}, \mathcal E_{g_m})$ can be regarded as the subgraph/substructure of $\mathcal G=(\mathcal V, \mathcal E)$ \textit{iff} there exist an injective function $\phi: \mathcal V_{g_m} \rightarrow \mathcal V$ s.t. $\{u,v\} \in \mathcal E_{g_m}$ and $\{\phi(u),\phi(v)\} \in \mathcal E$.
\end{definition}

\begin{definition}
\textbf{(Graph-level Learning)}: 
Graph-level learning takes a graph dataset $\mathbb G = \{\mathcal G_1,...,\mathcal G_N\}$ consisting of $N$ graphs as inputs and returns a function $f(\cdot)$ which maps a graph $\mathcal G_i$ to some output $f(\mathcal G_i)$. For instance, in the graph classification task, for any graph $\mathcal G_i'$ isomorphic to $\mathcal G_i$, we have $f(\mathcal G_i) = f(\mathcal G_i')$.
In other words, $f(\cdot)$ is permutation-invariant\footnote{The prediction results of a graph-level learning algorithm are invariant to any permutations of the order of nodes and/or edges of each input graph.}.

\end{definition}


%% file: 3-Development_and_Taxonomy.tex
\section{Taxonomy of Graph-level Learning Techniques}

This section provides a taxonomy of graph-level learning techniques.
Its categories include traditional learning, graph-level deep neural networks (GL-DNNs), graph-level graph neural networks (GL-GNNs), and graph pooling.
Each category is briefly introduced next. 
The taxonomy tree describing these four branches of graph-level learning with selected algorithms can be found in Fig. \ref{tree} in Appendix \ref{ap_tax}.\\
\textbf{Traditional Learning.}
As the historically dominant technique, traditional learning tries to solve the fundamental problem that is lacking feature representations of graphs, by manually defined features.
Given well-designed features (e.g., random walk sequences \cite{gartner2003graph}, 
frequently occurring substructure \cite{inokuchi2000apriori}), off-the-shelf machine learning models were used to tackle graph classification tasks, in a non-end-to-end fashion.
The form of traditional learning is less applicable to reality complex networks due to the computational bottlenecks, yet, it still provides great valuable insights, such as better interpretability and better ability to model irregular structures \cite{du2019graph}.\\
\textbf{Graph-Level Deep Neural Networks (GL-DNNs).} Towards the deep learning era, neural networks achieved wide success in representing Euclidean data (e.g., images and texts).
Thus, researchers try to apply deep neural networks to graph data, the tentative explorations include Skip-gram, RNNs, CNNs, and CapsNet.
These four types of deep neural networks were not initially designed to learn non-Euclidean data like graphs. 
Hence, one of the important issues with GL-DNNs is how to enable these deep neural networks to learn graph-structured data that varies in size and has irregular neighborhood structures.\\ 
\textbf{Graph-Level Graph Neural Networks (GL-GNNs).} GL-GNNs use graph convolution operations specifically proposed for graphs as the backbone for performing graph-level learning \cite{wu2020comprehensive}. 
Most GL-GNNs use the graph convolutions MPNNs frameworks because they are simple, easy to understand, and have linear complexity \cite{gilmer2017neural}. 
GL-GNNs condense the most fruitful achievements of graph-level learning. 
In addition, some practitioners integrate the advantages of MPNN-based GL-GNNs with other techniques, particularly traditional learning techniques, to improve graph-level learning.\\ 
\textbf{Graph Pooling.} GL-DNNs and GL-GNNs always encode graph information into node representations that cannot be directly applied to graph-level tasks, graph pooling fills this gap.
Graph pooling is a kind of graph downsizing technology where compact representations of a graph are produced by compressing a series of nodes into a super node \cite{zhang2020deep,liu2022graph}.
It is worthy to be recorded as a significant graph-level technique, as it is unique for graph-level learning without appearing in node-level and edge-level tasks.
In addition, graph pooling has great power to preserve more information (e.g., hierarchical structure) for graph-level tasks, resulting in an abundant literature of related methods.

%% file: 4-Traditional_Learning_Techniques.tex
\section{Traditional Learning}\label{TL}

Traditional graph-level learning algorithms work in a deterministic way, encoding graphs using handcrafted features.
Traditional graph-level learning methods can be divided into three main types: i.e., those based on graph kernels (GKs, Section \ref{41}), subgraph mining (Section \ref{42}), and graph embedding (Section \ref{43}). 
We summarize all discussed traditional graph-level learning models in Table \ref{table_traditional} in Appendix \ref{ap_tr}. 

\begin{figure*}[htbp!]
    \centering
    \includegraphics[width=1\textwidth]{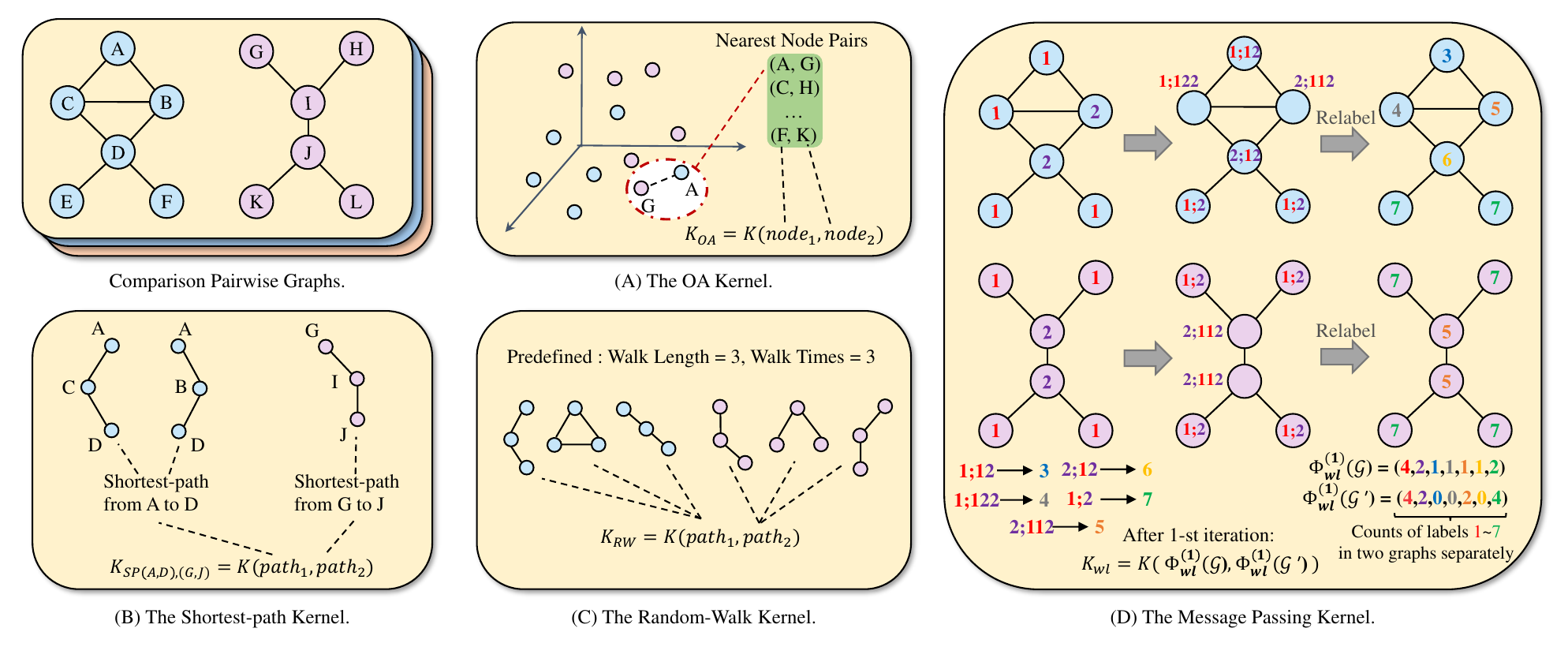}
    \caption{Different mechanisms of four graph kernels in decomposing and comparing pairwise graphs.}
    \label{fgraphkernel}
\end{figure*}

\subsection{Graph Kernels (GKs)}\label{41}

GKs perform graph-level learning based on kernel values (i.e., pair-wise graph similarities) \cite{}. 
Given a graph dataset $\mathbb G$, GKs decompose each graph $\mathcal G$ into a bag-of-graphs $S^{\mathcal G}=\{g_1,...,g_I\}$, where $g_i\subseteq \mathcal G$ and $g_i$ can be a node or a subgraph. 
Most GKs are based on the paradigm of an $R$-Convolution kernel \cite{haussler1999convolution} that obtains the kernel value $K_{R-conv} \left(\mathcal G,\mathcal G'\right)$ of two graphs $\mathcal G$ and $\mathcal G'$ by:
\begin{equation}
    K_{R-conv} \left(\mathcal G,\mathcal G'\right) = \sum_{i=1}^{I}\sum_{j=1}^{J} K_{parts}\left(g_i,g'_j\right),
\end{equation}
where $K_{parts}\left(g_i,g'_j\right)$ is the kernel function that defines how to measure the similarity between $g_i$ and $g'_j$. 
A kernel matrix that packages all kernel values is then fed into an off-the-shelf machine learning model, such as a support vector machine (SVM), to classify the graphs.

\subsubsection{Message Passing Kernels (MPKs)}\label{MPKs_formal} MPKs perform message passing on neighborhood structures to obtain graph representations.
The 1-dimensional Weisfeiler-Lehman (1-WL) algorithm\footnote{1-WL is also a well-known algorithm for graph isomorphism test.} \cite{weisfeiler1968reduction,shervashidze2011weisfeiler} is one of the most representative MPKs. 1-WL updates a node's label (or color) iteratively. An illustration of $1$-th iteration is shown in Fig. \ref{fgraphkernel} (D).
At the $h$-th iteration, 1-WL aggregates node $v$'s label $l^{(h-1)}(v)$ and its neighbor's labels $l^{(h-1)}(u), u\in\mathcal N(v)$ to form a multi-set\footnote{In a multiset, multiple elements are allowed to be the same instance.} of labels $\{l^{(h-1)}(v),sort(l^{(h-1)}(u):u\in\mathcal N(v))\}$.
Subsequently, 1-WL employs an injective hash function $\phi(\cdot)$ to map the $\{l^{(h-1)}(v),sort(l^{(h-1)}(u):u\in\mathcal N(v))\}$ into a new label $l^{(h)}(v)$. Formally:
\begin{equation}\label{MPK}
    l^{(h)}(v) = \phi\left(l^{(h-1)}(v),sort(l^{(h-1)}(u):u\in\mathcal N(v))\right).
\end{equation}
When $\phi(\cdot)$ no longer changes the labels of any nodes, 1-WL stops iterating and generates a vector $\phi_{wl}(\mathcal G)$ that describes $\mathcal G$. That is, 
\begin{equation}
    \phi_{wl}\left(\mathcal G\right) = [c^{(0)}(l^{(0)}_1),.., c^{(0)}(l^{(0)}_{I_0});...;c^{(H)}(l^{(H)}_1),...,c^{(H)}(l^{(H)}_{I_H})],
\end{equation}
where $l^{(h)}_i$ is the $i$-th label generated at the $h$-th iteration, and $c^{(h)}(l^{(h)}_i)$ counts the occurrences of nodes labeled with $l^{(h)}_i$ in the $h$-th iteration.
The kernel value of 1-WL between $\mathcal G$ and $\mathcal G'$ is the inner product of $\phi_{wl}\left(\mathcal G\right)$ and $\phi_{wl}\left(\mathcal G'\right)$:
\begin{equation}
    K_{WL}\left(\mathcal G,\mathcal G'\right) = <\phi_{wl}\left(\mathcal G\right),\phi_{wl}\left(\mathcal G'\right)>.
\end{equation}

The followed upgrading of 1-WL mainly focuses on aggregation and relabeling steps.
Hido and Kashima \cite{hido2009linear} replaced the hash function with a binary arithmetic giving rise to a faster $\phi(\cdot)$.
Morris \textit{et al.} \cite{morris2017glocalized} used the idea of $k$-WL to relabel node groups consisting of $k$ nodes that could form a connected graph. 
Theoretically, $k$-WL is more powerful than $1$-WL for distinguishing between graph structures. 
Further, Neumann \textit{et al.} \cite{neumann2016propagation} proposed a random label aggregation process based on node label distributions that only considers labels of part of neighbors. 
Random label aggregation saves time and computational resources making work on large-scale graphs more efficient.
Persistent Weisfeiler–Lehman (P-WL) \cite{rieck2019persistent} is the recent enhancement to MPKs that adds weighted edges into the aggregation process. To calculate the edge weight, P-WL measures the distance between the continuous iterative updated labels of two end nodes.
Additionally, P-WL can track changes in substructures that cannot be identified by 1-WL, such as cycles.

\subsubsection{Shortest-path Kernels (SPKs)}

SPKs denote the kernel value as a comparison between pair-wise node sequences (see Fig. \ref{fgraphkernel} B). For example, the shortest-path kernel \cite{borgwardt2005shortest} determines the shortest path between the vertices $v$ and $u$ via the Floyd-Warshall \cite{floyd1962algorithm} or Dijkstra's \cite{dijkstra1959note} algorithms. The distance between the pairwise shortest paths from $\mathcal G$ and $\mathcal G'$ is defined as the kernel value between them. Formally, 
\begin{equation}\label{shortestpath}
\begin{aligned}
    &K_{SP}\left(\mathcal G,\mathcal G'\right) = \sum\limits_{\substack{v,u \in V\\ v \neq u}} \sum\limits_{\substack{v',u' \in V'\\ v' \neq u' }} K_{Parts}\left(\left(v,u\right),\left(v',u'\right)\right) := 
    \begin{cases}
    K_D\left(P\left(v,u\right),P\left(v',u'\right)\right) & \text{if} \ l\left(v\right) \equiv l\left(v'\right) \land l\left(u\right) \equiv l\left(u'\right),\\
    0 & \text{otherwise},
    \end{cases}
\end{aligned}
\end{equation}

where $l(v)$ is the label of node $v$, $P\left(v,u\right)$ is the length of shortest path between vertices $v$ and $u$, and $K_D(\cdot, 
\cdot) $ is a kernel comparing the shortest path lengths.  
Nikolentzos \cite{nikolentzos2017shortest} proposed a variant of SPKs that draws on more information in a shortest path, such as node and edge labels, to calculate the distance of any two paths. 

\subsubsection{Random Walk Kernels (RWKs)}\label{RWKs_formal}
RWKs are another kernel method guided by node sequences. G{\"a}rtner \textit{et al.} \cite{gartner2003graph} was the first to propose a random walk kernel. This technique counts the same random walk sequences that pair-wise graphs both own. Performing random walks on $\mathcal G = (\mathcal V, \mathcal E)$ and $\mathcal G' = (\mathcal V', \mathcal E')$ simultaneously is the same as conducting random walks on a direct product graph $\mathcal G_{\times}  = (\mathcal V_{\times}, \mathcal E_{\times})$, where
\begin{equation}
\begin{aligned}
    \mathcal V_{\times} = \{\left(v,v'\right):v\in\mathcal V \land v'\in\mathcal V' \land l(v) \equiv l(v')\}, \quad \mathcal E_{\times} = \{\{\left(v,v'\right),\left(u,u'\right) \in \mathcal V_{\times}\}:\mathcal E_{v,u} \in \mathcal E \land \mathcal E'_{v',u'} \in \mathcal E' \}.
\end{aligned}
\end{equation}
Given $\mathcal G_{\times}$, the kernel function is defined as:
\begin{equation}\label{krw}
    K_{RW}\left(\mathcal G,\mathcal G'\right) = \sum\limits_{i=1}^{|\mathcal V_{\times}|}\sum\limits_{j=1}^{|\mathcal V_{\times}|}\left[\sum\limits_{p=0}^{P}\lambda_p \mathbf A_{\times}^p\right]_{ij},
\end{equation}
where $\mathbf A_{\times}$ is the adjacency matrix of $\mathcal G_{\times}$, 
$P$ is the predefined max length of random walking sequences, and $\lambda_p$ are the weights given to different $P$. 
$K_{RW}\left(\mathcal G,\mathcal G'\right)$ counts the occurrences of common walk paths in $\mathcal G$ and $\mathcal G'$ with lengths equal to or less than $P$. 

The random walk kernel in Eq. (\ref{krw}) assumes a uniform distribution for the beginning and ending probabilities of the walks across two graphs. 
However, Vishwanathan \textit{et al.} \cite{vishwanathan2010graph} proposed a generalized version of RWKs.  Specifically, they defined $\mathbf p$ and $\mathbf q$ as the beginning and ending probability vectors in $\mathcal G$, respectively. 
In addition, they used the Kronecker product operation $\otimes$ to derive $\mathbf A_{\times}$, that is $\mathbf A_{\times}= \mathbf A \otimes \mathbf A'$. Formally, the kernel value is:
\begin{equation}
    K_{RW}\left(\mathcal G,\mathcal G'\right) = \sum_{l=0}^{\infty} \mu_{l} (\mathbf q \otimes \mathbf q' \mathbf)^{\top} (\mathbf A_{\times})^l   (\mathbf p \otimes \mathbf p'),
\end{equation}
where $\mu_{l}$ is the convergence coefficient. 

RWKs suffer from a problem called tottering, where a random walk sequence traverses $v$ to $u$ and immediately returns to $v$ via the same edge.
To address tottering, Mah{\'e} \textit{et al.} \cite{mahe2004extensions} employed a second-order Markov random walk that considers the last two steps in the current random walk sequence when deciding the next step.

\subsubsection{Optimal Assignment Kernels (OAKs)} 

Fr{\"o}hlich \textit{et al.} \cite{frohlich2005optimal} was the first to propose OAKs. OAKs consider nodes as a basic unit for measuring kernel values. Of all the GKs introduced in this paper, OAKs are the only family of GKs that do not belong to $R$-Convolution paradigm. Specifically, given a fixed $i$ in Eq. (1), OAKs only add in the maximum similarity value between $g_i$ and $g'_j$ where $j\in\{1,...,J\}$. Formally, OAKs are defined as:

\begin{equation}\label{OA}
    K_{OA} \left(\mathcal G,\mathcal G'\right) = \begin{cases}
    \max\limits_{\pi \in \prod_J}  \sum_{i=1}^{I} K_{parts}\left(g_i,g'_{\pi[i]}\right), & \text{if} \ J \geq I\\
    \max\limits_{\pi \in \prod_I}  \sum_{j=1}^{J} K_{parts}\left(g_{\pi[j]},g'_j\right),& \text{otherwise}
    \end{cases}
\end{equation}
where $\prod_I$ represents all permutations of the indexes of a bag-of-graphs $\{1,...,I\}$, and $\pi$ is the optimal node permutation to reach maximum similarity value between two graphs.

Searching for a pair-wise element with the maximum similarity tends to be a highly time-consuming process.
Hence, to reduce the time requirement of this task, Johansson et al. \cite{johansson2015learning} mapped the graphs in geometric space and then calculated the Euclidean distance between pair-wise nodes. 
This method enables OAKs to use approximate nearest neighbors algorithms in Euclidean space as a way to speed up the process. 
Transitive Assignment Kernels (TAKs) \cite{schiavinato2015transitive,pachauri2013solving} are variants of OAKs. 
Unlike OAKs that search for the optimal assignment among pair-wise graphs, TAKs identify node permutations that with the most similar node pairs among three or more graphs. 
OAKs have been confined to node similarity measurement, although they can be extended to measure subgraph similarities so as to capture a graph's topological information \cite{woznica2010adaptive}. As discussed next, we introduce the GKs with subgraph information.

\subsubsection{Subgraph Kernels (SGKs)} 
SGKs calculate the similarity between two graphs by comparing their subgraphs.
For example, the representative SGK \textemdash Graphlet Kernel \cite{shervashidze2009efficient} uses either depth-first search (DFS) or sampling to identify the subgraphs. 
With these subgraphs, the vector $\phi_{SG}\left(\mathcal G\right) = [c^{\left(\mathcal G\right)}_{\mathcal T_1},...,c^{\left(\mathcal G\right)}_{\mathcal T_N}]$ is then used to describe the graph $\mathcal G$, where $\mathcal T_i$ means the $i$-th isomorphism type of subgraphs, $N$ is the total number of subgraphs' types, and $c^{\left(\mathcal G\right)}_{\mathcal T_i}$ counts the occurrences of the $\mathcal T_i$ category subgraphs in graph $\mathcal G$.
Graphlet's kernel value is then defined as the inner product of $\phi_{SG}\left(\mathcal G\right)$ and $\phi_{SG}\left(\mathcal G'\right)$:
\begin{equation}
    K_{SG}\left(\mathcal G,\mathcal G'\right) = <\phi_{SG}\left(\mathcal G\right),\phi_{SG}\left(\mathcal G'\right)>.
\end{equation}

There are several different implementations of SGKs kernel functions. 
For instance, Wale \textit{et al.} \cite{wale2008comparison} employed a min-max kernel $\displaystyle{ \tfrac{\sum_{i=1}^N min(c^{\left(\mathcal G\right)}_{\mathcal T_i},c^{\left(\mathcal G'\right)}_{\mathcal T_i})}{\sum_{i=1}^N max(c^{\left(\mathcal G\right)}_{\mathcal T_i},c^{\left(\mathcal G'\right)}_{\mathcal T_i})}}$ to measure the distance between two graphs.
Subgraph Matching Kernels (SMKs) \cite{10.5555/3042573.3042614} calculate the similarity between two subgraphs by counting the number of nodes with the same labels.
Then the similarities between all pairwise subgraphs sourced from the two graphs are summed as the kernel value of the SMKs.
Methods of identifying the subgraphs in SGKs have also been explored. 
For example, Neighborhood Subgraph Pairwise Distance Kernels (NSPDK) \cite{costa2010fast} denotes the subgraphs as the first-, second-, and third-hop neighborhoods of pairwise vertices with the shortest path of a predefined length. 
However, the main contributions of SGKs lie in assessing the similarity of graphs in terms of a set of selected subgraphs, not how the subgraphs are chosen.
More detailed and sophisticated subgraph mining methods are demonstrated next.

\subsection{Subgraph Mining}\label{42}

\begin{figure*}[htbp!]
\begin{minipage}{0.48\linewidth}
\vspace{3pt}
\centerline{\includegraphics[width=\textwidth]{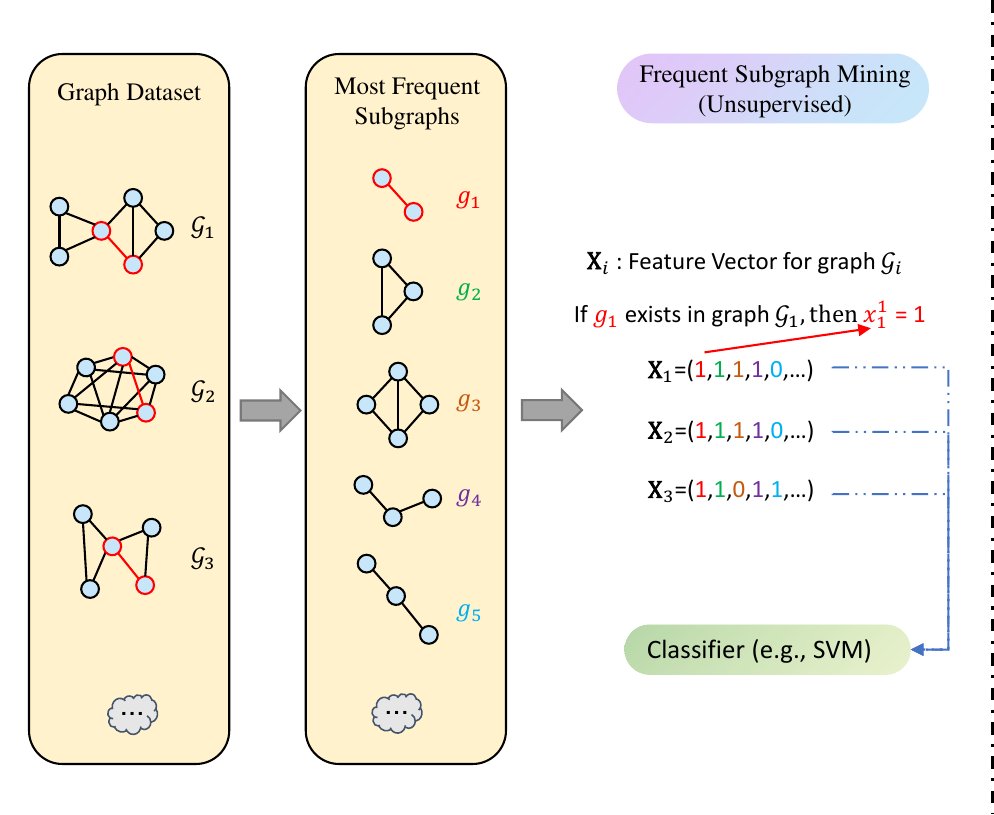}}
\centerline{(A) Frequent Subgraph Mining (FSM).}
\end{minipage}
\begin{minipage}{0.48\linewidth}
\vspace{3pt}
\centerline{\includegraphics[width=\textwidth]{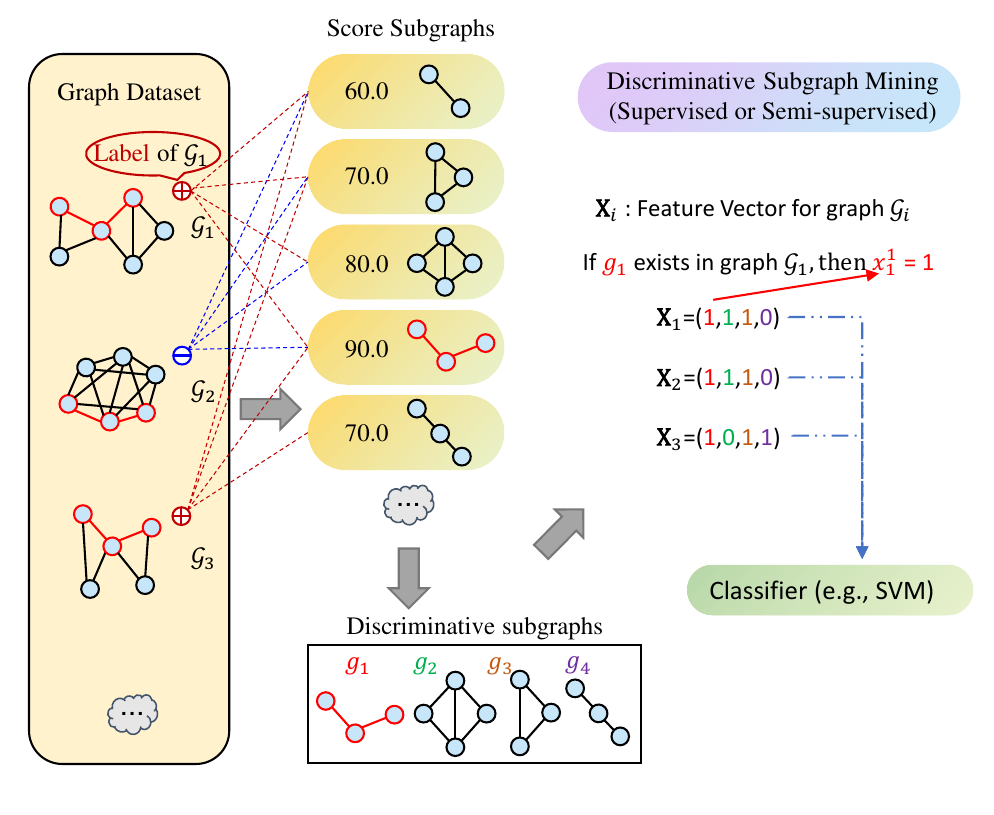}}
\centerline{(B) Discriminative Subgraph Mining (DSM).}
\end{minipage}
\caption{Different subgraph extraction methods of FSM and DSM.}    \label{fsubgraph}
\end{figure*}

Subgraph mining is similar to SGKs, where the vector $\textbf x_i = [x_i^1,...,x_i^M]^\top$ is taken as a graph-level representation of the graph $\mathcal G_i$. 
Here, $x_i^m\in\{0,1\}$, $x_i^m=1$ if $g_m \subseteq \mathcal G_i$, otherwise, $x_i^m=0$. The established graph-level representation is then directly input into an off-the-shelf machine learning model, such as SVM classifier, for downstream tasks. 
What is different about subgraph mining algorithms is that they place particular emphasis on how to extract the optimal subgraph set $\mathcal S^* = \{g_1,...g_T\}$ from the subgraph set $\{g_1,...g_M\}$, where $g_1, ...,g_M$ denote all possible subgraphs of $\mathbb G =\{\mathcal G_1,...,\mathcal G_N\}$. Techniques for extracting subgraphs can be divided into two branches depending on how the supervision information is used.
Frequent subgraph mining is the unsupervised method, as illustrated in Fig. \ref{fsubgraph} A, while discriminative subgraph mining is the supervised or semi-supervised method (see Fig. \ref{fsubgraph} B).

\subsubsection{Frequent Subgraph Mining (FSM)}\label{FSM} FSM identifies the subgraphs whose frequency of occurence in $\mathbb G$ sits over a predefined threshold $\delta$.
These subgraphs are then added to $\mathcal S^*$. 
Apriori-like algorithms, such as AGM \cite{inokuchi2000apriori} and FSG \cite{kuramochi2001frequent}, enumerate subgraphs from size one to a predefined largest size as candidates for $\mathcal S^*$. 
In the enumeration, these apriori-like algorithms pick up the candidates that occur more frequently than $\delta$ and add them to $\mathcal S^*$. 
Others subgraphs are dropped and expansions based on those subgraphs are no longer considered.
Testing for subgraph isomorphism with vast numbers of candidate subgraphs can mean apriori-like algorithms suffer from computation bottlenecks. 
To address this issue, gSpan \cite{yan2002gspan} employs a depth-first-search (DFS) strategy to search subgraphs, while assigning a unique DFS code of minimum length for each subgraph searched. 
gSpan can then do a quick check for isomorphism by simply comparing the DFS codes of pairwise subgraphs.

\subsubsection {Discriminative Subgraph Mining (DSM)} DSM extracts discriminative subgraphs from a set of all possible subgraphs of $\mathbb G$ based on the label information. 
Given a binary graph classification task, Thoma \textit{et al.} \cite{thoma2009near} defined an evaluation criterion called CORK which describes the discriminative score of a subgraph set $\mathcal S$, $\mathcal S \subseteq \{g_1,...,g_M\}$. Formally,
\begin{equation}
\begin{aligned}
    CORK(\mathcal S) = -1 \times num (\mathcal G_i, \mathcal G_j), \quad s.t. \ \mathcal G_i \subset \mathbb G_{+} \ \land  \ \mathcal G_j \subset \mathbb G_{-} \ \land \ \forall g_m \in \mathcal S : x_i^m = x_j^m,
\end{aligned}
\end{equation}
where $num(\cdot)$ counts the number of pairs of graphs $(\mathcal G_i, \mathcal G_j)$ satisfying the specific conditions. 
$\mathbb G_{+}$ is the set of graphs with positive labels, while $\mathbb G_{-}$ is the set of graphs with negative labels.
The optimal subgraph set $\mathcal S^*$ has the highest CORK score among all possible subgraph sets $\mathcal S$ containing $T$ subgraphs, denoted as:
\begin{equation}
    \mathcal S^* = \mathop{argmax}\limits_{\mathcal S \subseteq \{g_1,...,g_M\}} CORK(\mathcal S)  \quad s.t. \ |\mathcal S| \leq T.
\end{equation}

The CORK score can also be used to prune the subgraph search space of gSpan, mentioned in Section \ref{FSM}. 
More specifically, if replacing any existing element of $\mathcal S^*$ with a subgraph $g_m$ does not $\mathcal S^*$'s CORK score, gSpan will no longer perform DFS along $g_m$. 
To speed up DSM based on discriminative scores and gSpan, Yan \textit{et al.} \cite{yan2008mining} proposed LEAP, which initializes an optimal subgraph set $\mathcal S^*$ with frequent subgraphs. 
In this way, LEAP prunes gSpan's search space right at the beginning. In addition, Kong \textit{et al.} \cite{kong2010multi} and Wu \textit{et al.} \cite{wu2014multi} expanded DSM to the multi-label\footnote{Each graph owns more than one label, such as a drug molecular can own different labels to represent anti-cancer effects for various cancers, e.g., breast cancer (+) and lung cancer (-).} and multi-view\footnote{An object has different views, where each view can represent a separate graph, e.g., a scientific publication network is shown as two graphs, an abstract graph demonstrating the keywords correlations in the abstract of papers, and a reference citation graph about citation relationships.} scenarios, respectively. 
Note, however, that all the DSM methods discussed are supervised methods.
In terms of semi-supervised subgraph mining, Kong and Yu \cite{kong2010semi} proposed gSSC which maps each graph into a new feature space by $\mathcal S^*$. Unlabeled graphs are separated from each other in the new feature space. 
In the labeled group, graphs with the same label are close, whereas graphs with different labels remain distant. In addition, Zhao \textit{et al.} \cite{zhao2011positive} only used the positively labeled graphs and unlabeled graphs to select $\mathcal S^*$ when performing binary graph classification tasks. This is because sometimes the real-world data is composed of an incomplete set of positive instances and unlabeled graphs.

\subsection{Non-learnable Graph Embedding}\label{43}

Graph embeddings are the compression of graphs into a set of lower-dimensional vectors.
Some non-learnable graph embedding methods extract graph-level representations from the inherent properties of graphs, e.g., their topologies and eigenspectrums. 

Local Degree Profile (LDP) \cite{cai2018simple} summarizes the degree information of each node and its 1-hop neighbors as node features. 
LDP constructs graph representations by building an empirical distribution or histogram of the hand-crafted node features.
In addition to node degree, non-learnable graph embedding can also leverage anonymous random walk sequences to describe a graph's topological information. 
Specifically, anonymous random walks record the status change of node labels. Two anonymous random walk sequences $A \to B \to A$ and $B \to A \to B$ can be both written as $1 \to 2 \to 1$. Anonymous Walk Embeddings (AWE) \cite{ivanov2018anonymous} encodes a graph via an $n$-dimensional vector in which each element represents the occurrence frequency of a specific anonymous random walk sequence.

In spectral graph theory \cite{chung1997spectral}, the spectrum of a graph is determined by its topology. Based on this theory, the Family of Graph Spectral Distances (FGSD) method \cite{verma2017hunt} proposes that the distance between the spectrums of two graphs can be used to test whether the graphs are isomorphic. 
Thus, the histogram of the spectrum is used to construct a graph-level representation. 
Analogously, A-DOGE \cite{sawlani2021fast} depicts a graph by computing the spectral density across its eigenspectrum. 
However, these methods are limited to use with small graphs given the prohibitive costs of computing eigenspectrum decompositions with large-scale graphs. 
As a possible solution to this limitation, SlaQ \cite{tsitsulin2020just} uses stochastic approximations as a way of quickly calculating the distance between two graphs' spectral densities. 
More specifically, these authors employed von Neumann graph entropy (VNGE) \cite{braunstein2006laplacian,chen2019fast} as a way of approximately representing the spectral properties of the graphs.
In turn, this approximation supports fast computation by tracing a Laplacian matrix of the graph. 
Liu \textit{et al.} \cite{liu2021bridging} proposed another fast approximation method involving VNGE, which is based on deriving the error bound of the approximation estimation.

%% file: 5-GLNN-DNN.tex
\section{Graph-Level Deep Neural Networks (GL-DNNs)}\label{GL-DNNs}

GL-DNNs form the basis of a pioneering set of works that employ deep learning techniques to achieve graph-level learning. Researchers have explored graph-level learning techniques based on classic deep neural networks including skip-gram neural network, recurrent neural networks (RNNs), convolution neural networks (CNNs), and capsule neural networks (CapsNets) to achieve Skip-gram-based (see Section \ref{skipgram}), RNN-based (see Section \ref{5.1}), CNN-based (see Section \ref{5.2}), and CapsNets-based (see Section \ref{5.3}) GL-DNNs, respectively. 
The representative GL-DNNs mentioned in this section are summarized in Table \ref{table_dnn} in Appendix \ref{ap_dnn}.

\subsection{Skip-gram-Based GL-DNNs}\label{skipgram} 

Skip-gram \cite{mikolov2013efficient} is a widely used unsupervised neural networks, to predict the context words for the target word.
Initially, the researchers built a skip-gram model based on the relationship between two adjacent subgraphs, namely subgraph2vec \cite{narayanan2016subgraph2vec}. 
Subgraph2vec first takes the ($d$-$1$)-, $d$-, ($d$+1)-hop neighborhoods of the $v$th selected node in the graph $\mathcal G_i$ as three subgraphs $g^i_{v-1}$, $g^i_{v}$, $g^i_{v+1}$, respectively, where $ d\ge1$ is a predefined value. $\{\mathbf w^1_{1-1},...,\mathbf w^1_{V+1};...;\mathbf w^N_{1-1},...,\mathbf w^N_{V+1}\}$ are the randomly initialized embeddings of all sampled subgraphs $\{g^1_{1-1},...,g^1_{V+1};...; g^N_{1-1},...,g^N_{V+1}\}$ respectively,  where $N$ represents the total number of graphs, and $V$ is the number of selected nodes in each graph. Then, the Skip-gram model is used to update the subgraph embeddings. 
The Skip-gram model takes $\mathbf w^i_{v}$ as its input, and predicts the context of $\mathbf w^i_{v}$ (i.e., $\mathbf w^i_{v-1}$ and $\mathbf w^i_{v+1}$).
Then the prediction results are back-propagated to update $\mathbf w^i_{v}$. 
To summarize, subgraph2vec's learning objective is to maximize the following log-likelihood:
\begin{equation}
\sum_{i=1}^{N} \sum_{v=1}^{V}\log \operatorname{Pr}\left(\mathbf w^i_{v-1}, \ldots, \mathbf w^i_{v+1} \mid \mathbf w^i_{v}\right).
\end{equation}

Another method, Graph2vec \cite{narayanan2017graph2vec} was designed to tackle graph representation tasks. 
By establishing a semantic association between a graph and its sampled subgraphs, Graph2Vec employs the idea of Skip-gram to learn a graph embedding. 
Following this work, Dang \textit{et al.} \cite{nguyen2018learning} replaced the sampled subgraphs in Graph2vec with frequent subgraphs that have more discriminative features for graph classification tasks.

\subsection{RNN-Based GL-DNNs}\label{5.1}

RNNs are particularly good at learning sequential data, such as text and speech.
There are two main types of algorithms that apply RNNs to graph-level learning. 
One type transforms graphs into sequential-structured data.
The other aggregates neighborhood information about the target node and relabels those aggregated features through an RNN.
This is similar to Message Passing Kernels (MPKs, Section \ref{MPKs_formal}).

A natural way to capture the sequential information in graphs is to use a series of random walk paths to represent a graph. 
For example, GAM \cite{lee2018graph} employs a long short-term memory (LSTM) model to guide a random walk on graphs. 
Meanwhile, the LSTM model generates a representation for the walk sequence to describe the graph.
In addition, Zhao \textit{et al.} \cite{zhao2018substructure} proposed an RNN-based graph classification algorithm called SAN. 
Starting from a node, SAN employs an RNN model that adds nodes and edges to form an informative substructure whose representation is progressively generated by the RNN model.
A graph-level representation that can be used for graph classification tasks is then generated by summing all the representations of the formed substructures.
Given a graph generation task, NetGAN \cite{bojchevski2018netgan} uses an LSTM model as a generator to yield fake walk sequences,
while a discriminator disambiguates the graph's real walk sequences from the generated fake ones to reverse-train the generator.
Another graph generation model Graphrnn \cite{you2018graphrnn} creates various permutations of graphs, with various combinations of nodes and edges as sequential data to be input into an RNN model. 

The second category of RNN-based GL-DNNs implements neural networks version of MPKs through RNN models.
As such, the algorithms in this category can be viewed as the predecessors of MPNNs.
Scarselli \textit{et al.} \cite{scarselli2008graph} recurrently updated node embeddings until reaching a stable situation, that is:
\begin{equation}
    \mathbf h_v^{(k)} = \sum_{u\in \mathcal N(v)} f_{w}
    \left( 
    \mathbf x_v, \mathbf s_{u, v}, \mathbf x_u, \mathbf h_u^{(k-1)}
    \right),   
\end{equation}
where $h_u^{0}$ is randomly initialized and $f_{w}(\cdot)$ is a parametric function that maps vectors into a concentrated space to shorten their distance. 
To address the graph-level task, a super node connected with all the other nodes is used to output the representation of the whole graph.
In addition, Li \textit{et al.} \cite{li2016gated} proposed the idea of using a gated recurrent unit (GRU) to relabel the aggregated information from the 1-hop neighborhoods of the center node.
This approach reduces the recurrent process for updating node embeddings to a fixed number of steps and avoids control convergence parameters, formulated as:
\begin{equation}\label{RNNMPK}
    \mathbf h_v^{(k)} = \text{GRU}\left(\mathbf h_v^{(k-1)}, \text{AGG}^{(k)} \left(\mathbf h_u^{(k-1)}: u\in \mathcal N(v)\right) \right),
\end{equation}
where $\mathbf h_v^{(k)}$ represents the node representation of $v$ at the $k$-th iteration, $\mathbf h_v^{(0)}$ is the node feature $\mathbf x_v$, and here $\text{AGG}$ is a weighted sum aggregation function.
This algorithm continues the recurrent process until it hits the predefined $K$ number of iterations needed to form the node representations. 
A graph-level representation is then produced via:
\begin{equation}\label{GRUPooling}
\mathbf{h}_{\mathcal{G}}=\tanh \left(\sum_{v \in \mathcal{V}} f_{t}\left(\mathbf{h}_{v}^{(K)}, \mathbf{h}_{v}^{(0)}\right) \odot \tanh \left(\mathbf{h}_{v}^{(K)}\right)\right),
\end{equation}
where $f_{t}\left(\cdot\right)$ is a softmax function guided by an attention mechanism, that preserves and aggregates valuable node representations for specific graph-level tasks. $\tanh\left(\cdot\right)$ is an activation function, and $\odot$ is element-wise multiplication.

\subsection{CNN-Based GL-DNNs}\label{5.2}

Another significant deep learning technique that works in the Euclidean domain is CNN. 
Here, grid-structured data, such as images, are studied.  
Similar to RNN-based GL-DNNs, there are two main branches of CNN-based graph-level learning.
In Appendix \ref{ap_cnn}, Fig. \ref{fGLDNN} depicts the details of these two different branches.

The first branch sorts nodes and arranges the node features to form a concentration matrix, of grid-structured data, to train the CNNs.
PATCHY-SAN \cite{niepert2016learning} selects a fixed number of neighbors of a central node and sorts neighbors to concatenate their features as the grid-structured feature matrix. By choosing a series of central nodes, PATCHY-SAN constructs some matched feature matrices. 
Finally, a graph-level representation is produced by the CNN model from the concatenation matrix of all built feature matrices. 
In addition, Kernel Convolutional Neural Network (KCNN) \cite{nikolentzos2018kernel} sorts all vertices in a graph to form grid-structured data. 
A neighborhood graph is built for each vertex and a kernel matrix is constructed by implementing the kernel function (i.e., an SPK or an MPK) between all pairwise neighborhood graphs. In this work, the grid-structured data for feeding up CNN is the kernel matrix, where each row is a vector describing the similarities between the neighborhood graph of the matched index vertex and the other neighborhood graphs.

The second branch involves CNN-guided neural network versions of MPKs.
These methods have two main steps: aggregating neighborhood information to the central node, and using the convolution operation to relabel the aggregated features.
NN4G \cite{micheli2009neural} performs a convolution kernel upon 1-hop neighbors for updating the center node and outputs the graph-level representations based on the node embeddings produced by each convolution layer, which is defined as:
\begin{equation}
    \mathbf h_v^{(k)} = f\left( 
    \mathbf w^{(k-1)^{\top}} \mathbf x_v + \sum_{i=1}^{k-1} \mathbf w_{k,i}^{\top} \sum_{u\in \mathcal N(v)} \mathbf h_u^{(i)}
    \right), 
    \qquad 
    \mathbf{h}_{\mathcal{G}} = f\left( 
    \sum_{k=1}^{K} \mathbf w_k \frac{1}{|\mathcal V|} \sum_{v \in \mathcal V} \mathbf h_v^{(k)}
    \right), 
\end{equation}
where $f(\cdot)$ is a linear or sigmoidal function and $h_v^{(0)}=0$.
Another related work, ECC \cite{simonovsky2017dynamic} concatenates 1-hop neighbor embeddings $\left(\mathbf h_u^{(k-1)}: u\in \mathcal N(v)\right)$ around the central node $v$ to construct a feature matrix by the $k$-th iteration. 
Subsequently, a convolution and average operation is executed on the aggregated neighbor feature matrix to obtain a representation for the central node.
Then a graph-level representation is produced via max-pooling the node embeddings. 
\begin{equation}
\begin{aligned}
    \mathbf H = [\mathbf h_u^{(k-1)}: u\in \mathcal N(v)], \quad  \mathbf h_v^{(k)} = \frac{1}{|\mathcal N(v)|} \left(\mathbf W \odot \mathbf H\right) + b^{(k)},\quad \mathbf{h}_{\mathcal{G}} = \text{MaxPooling}\left(\mathbf h_v^{(K)} :v\in\mathcal V \right).
\end{aligned}
\end{equation}

 Moreover, Diffusion CNN (DCNN) \cite{NIPS2016_390e9825} aggregates multi-hop neighborhood features to the central nodes through a matrix multiplication $\mathbf P \mathbf X$, where $\mathbf P = [\mathbf A, \mathbf A^2, ..., \mathbf A^h] \in \mathbb R^{h \times n\times n}$ is a three-dimensional tensor containing multi-hop (i.e., 1-, 2-, ..., h-hops) adjacent matrices and  $\mathbf X \in \mathbb R^{n\times f}$ is the node features matrix. $\mathbf P \mathbf X \in \mathbb R^{h \times n\times f}$ represents the updated node features after multi-hop aggregation. 
For graph classification tasks, DCNN permutes the dimensions giving $\mathbf P \mathbf X \in \mathbb R^{n \times h\times f}$ and all node representations are averaged as $\mathbf P^{*} \in \mathbb R^{h\times f}$. 
Subsequently, a convolution operation is implemented on $\mathbf P^{*}$ to produce a graph-level representation. The convolution operation can be defined as
$
    \mathbf{h}_{\mathcal{G}} = f\left(\mathbf W \odot \mathbf P^{*}\right),
$
where $f\left(\cdot\right)$ is a nonlinear activation function, and $\mathbf W$ is a trainable weight matrix for convolution and summation.

\subsection{CapsNet-Based GL-DNNs}\label{5.3}

CapsNets \cite{hinton2011transforming} were originally designed to capture more spatial relationships between the partitions of an entity than CNNs. CapsNets are available to assemble vectorized representations of different features (e.g., colors, textures) to a capsule dealt with by a specific network. 
Thus, applying a CapsNet to a graph preserves rich features and/or structure information at the graph level.

Graph Capsule Convolutional Neural Networks (GCAPS-CNN) \cite{verma2018graph} iteratively aggregates neighbor information under different statistical moments (e.g., mean, standard deviation) to form a capsule representation of the central node, formulated as:
\begin{equation}
\mathbf h_v^{(k)} = \frac{1}{\left|\mathcal N(v)\right|}
\left[\begin{array}{cc}
 \left(\sum\limits_{u\in \mathcal N(v)}  \mathbf  h_u^{(k-1)}\right) \mathbf W_1 \text { (mean) } \\
 \left(\sum\limits_{u\in \mathcal N(v)} \left(\mathbf  h_u^{(k-1)}-\mu\right)^{2}\right) \mathbf W_2   \text { (std) } \\
\left( \sum\limits_{u\in \mathcal N(v)} \left(\frac{\mathbf  h_u^{(k-1)}-\mu}{\sigma}\right)^{3}\right) \mathbf W_3   \text { (skewness) } \\
\vdots 
\end{array}\right],
\end{equation}
where $\left(\mathbf W_1,\mathbf W_2,...\right)$ are the learnable weight matrices for mapping the aggregated features into a uniform hidden space with a dimensionality of $h$. 
If the number of statistical moments is $p$ and the final iteration number is $K$, each node will be represented as $\mathbf h_v^{(K)} \in \mathbb R^{p \times h}$, and the matrix of all $n$ node embeddings will be $H^{(K)} \in \mathbb R^{n \times p \times h}$. 
This approach employs a covariance function as the permutation-invariant layer to output a graph-level representation, defined as:
\begin{equation}
    \mathbf{h}_{\mathcal{G}} = \frac{1}{n} (H^{(K)}-\mu)^{\top}(H^{(K)}-\mu).
\end{equation}
CapsGNN \cite{xinyi2018capsule} iteratively aggregates node features to a center node, and, in turn, adds the aggregation results of each iteration to a capsule representation of the central node. 
An attention mechanism is then applied to all node capsules so as to generate a graph capsule that can be plugged into a capsule network for graph classification.
Mallea \textit{et al.} \cite{mallea2019capsule} employs the same approach as PATCHY-SAN \cite{niepert2016learning} to find substructures in graphs, while the feature matrices of searched substructures are assembled in a capsule network for graph classification.

%% file: 6-GLNN-GNN.tex
\section{Graph-Level Graph Neural Networks (GL-GNNs)}\label{GL-GNNs}

This section focuses on GL-GNNs, which are the most influential graph-level learning techniques at present. 
The cornerstone branch of GL-GNNs \textemdash Message Passing Neural Networks (MPNNs) (see Section \ref{MPNN_INTRO}) \textemdash are introduced first, followed by. 
Some emerging methods in GL-GNNs, such as subgraph-based methods (see Section \ref{subgraph-based GL-GNNs}) and graph kernel-based methods (see Section \ref{kernel-based GL-GNNs}). 
Notably, these emerging approaches take advantage of some of the insights from traditional graph-level learning methods. 
In addition, we review progress in spectral GL-GNNs (see Section \ref{spectralgnn}), which push graph-level learning forward through spectrum properties. 
There are also some contents related to GL-GNNs in Appendix \ref{ap_gnn}, such as contrastive learning-based approaches (see Appendix \ref{contrastive GL-GNNs}), 
the expressivity (see Appendix \ref{expressivity}), generalizability (see Appendix \ref{Generalizability}), and explainability (see Appendix \ref{Explanation-GNN}) of GL-GNNs. 
Please refer to Table \ref{table_gnn} in Appendix \ref{ap_gnn} for the GL-GNNs discussed in this section.

\subsection{Message Passing Neural Networks (MPNNs)}\label{MPNN_INTRO}

\begin{figure*}[htp!]
    \centering
    \includegraphics[width=1\textwidth]{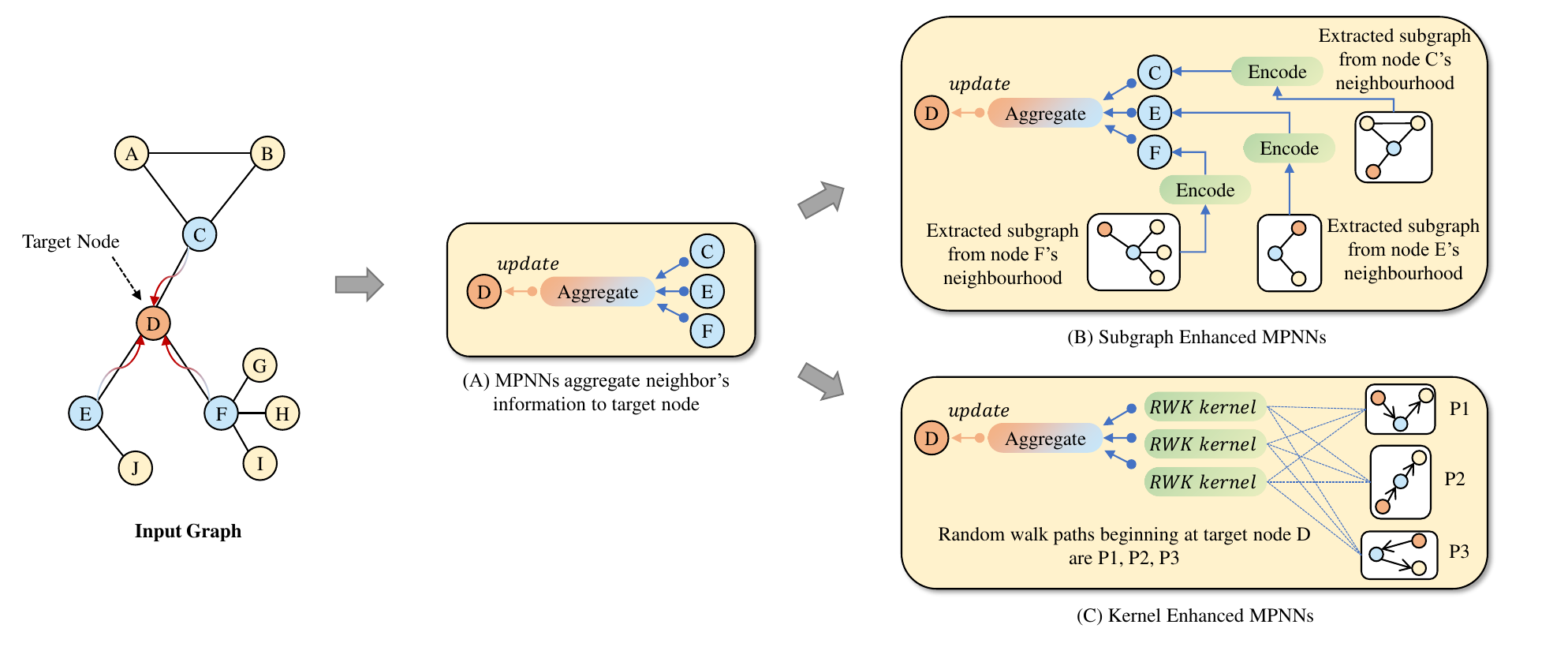}
    \caption{Different mechanisms of MPNNs, Subgraph Enhanced MPNNs, and Kernel Enhanced MPNNs. In Subgraph Enhanced MPNN, we used 1-hop neighborhoods as the subgraph for easy understanding, but the specific subgraph extraction is up to the article.}
    \label{fmpnn}
\end{figure*}

As mentioned, researchers have developed RNN- and CNN- based versions of MPKs. 
However, as the influence of deep learning has expanded, researchers have also developed various feedforward versions of Message Passing Kernels (MPKs, refer to Section \ref{MPKs_formal}). 
Collectively, these are called MPNNs.
MPNNs are similar to RNN-based MPKs in Eq. (\ref{RNNMPK}), but MPNNs set different weights in separate layers rather than sharing weights in all layers.
Gilmer \textit{et al.} \cite{gilmer2017neural} summarizes a collection of MPNNs \cite{battaglia2016interaction,schutt2017quantum,duvenaud2015convolutional} and further proposes a unified framework for this branch of techniques, as shown in Fig. \ref{fmpnn} (A) and denoted as:
\begin{equation}
    \mathbf h_v^{(k)} = U^{(k)}\left(\mathbf h_v^{(k-1)},\sum\limits_{u \in \mathcal{N}(v)} M^{(k)}\left(\mathbf h_v^{(k-1)},\mathbf h_u^{(k-1)},\mathcal E_{v,u}\right)\right),
\end{equation}
where $\mathbf h_v^{(0)} = \mathbf x_v$, 
$M^{(k)}$ is a function that outputs the passed message for the target node based on itself and its neighbors, and $U^{(k)}\left(\cdot\right)$ updates the embedding of the target node.
After multiple iterations, the node embeddings $\mathbf h_v^{(k)}$ learn the local structure information and the graph-level topology has distributed in all nodes. 
A readout function reads all node embeddings and outputs a graph-level representation, that is:
\begin{equation}
    \mathbf h_{\mathcal G} = \text{readout}\left(\mathbf h_v^{(k)}:v\in\mathcal V\right).
\end{equation}

MPNNs have become the mainstream of graph-level studies \cite{gilmer2017neural}.
They are also representative of spatial-based GL-GNNs since they are easy to use through matrix operations.
Lastly, the time and memory complexity of MPNNs only grows linearly with the graph size, making this a very practical approach for large sparse graphs.
In recent years, practitioners have developed numerous enhanced versions of MPNNs, including subgraph-enhanced MPNNs (see Section \ref{subgraph-based GL-GNNs}), and kernel-enhanced MPNNs (see Section \ref{kernel-based GL-GNNs}).
\subsection{Subgraph-Based GL-GNNs}\label{subgraph-based GL-GNNs}

In recent years, investigating GL-GNNs that are capable of capturing more topological information has been a crucial stream of study. This is especially, since a number of works have uncovered structure-aware flaws in MPNNs. 
To this end, practitioners have devised subgraph-based GL-GNNs, which leverage the rich structural information in subgraphs. 
These subgraph-based GL-GNNs can be divided into two branches.
The first branch enhances an MPNN by injecting the subgraph information into the aggregation process, as outlined in Fig. \ref{fmpnn} (B).
The other branch borrows the graphlet idea and decomposes the graph into a few subgraphs, merging multiple subgraph embeddings to produce an embedding of the entire graph.

\subsubsection{Subgraph Enhanced MPNNs} As mentioned, MPNNs learn topological information via a neighborhood aggregation process. 
However, standard MPNNs only aggregate node features, not structural information.
Therefore, a straightforward way of strengthening an MPNNs is to enrich the features of the nodes or edges with subgraph information.
Graph Substructure Network (GSN) \cite{bouritsas2022improving}, for example, counts the number of occurrences of a predefined subgraph pattern $g_1,...,g_M$ (e.g., a cycle or a triangle) that involves the target node $v$ or edge $\mathcal E_{v,u}$.
From these, subgraph feature vectors are constructed for $v$ as $\mathbf x_{v}^g$ or for $\mathcal E_{v,u}$ as $\mathbf S_{v,u}^g$, denoted as:
\begin{equation}
\begin{cases}
    x_v^{g_m} = |\{g_s \simeq g_m: v \in \mathcal V, v \in \mathcal V_{g_s}, g_s \subseteq \mathcal G\}|,\quad \mathbf x_{v}^g = [x_v^{g_1},...,x_v^{g_M}]^{\top} \ \textbf{(Node)},\\ 
     S_{v,u}^{g_m} = |\{g_s \simeq g_m: \mathcal E_{v,u} \in \mathcal E, \mathcal E_{v,u} \in \mathcal E_{g_s},g_s \subseteq \mathcal G\}|,  \quad \mathbf S_{v,u}^g = [S_{v,u}^{g_1},...,S_{v,u}^{g_M}]^{\top} \ \textbf{(Edge)},
\end{cases} 
\end{equation}
where $g_m$ is a predefined subgraph pattern, and $g_s \simeq g_m$ means $g_s$ is isomorphic to $g_m$, $x_v^{g_m}$ counts the number of isomorphic subgraphs $g_s$ containing the node $v$, and $S_{v,u}^{g_m}$ indicates the number of isomorphic subgraphs $g_s$ containing the edge $\mathcal E_{v,u}$. 
As a last step, the subgraph feature vectors for the node $\mathbf x_{v}^g$ and the edge $\mathbf S_{v,u}^g$ are injected into the aggregation layer, which is defined as:
\begin{equation}
\begin{aligned}
\mathbf h_v^{(k)} =  U^{(k)}\left(\mathbf h_v^{(k-1)},\mathbf m_v^{(k)}\right), \quad \mathbf m_v^{(k)} = \begin{cases}
     \sum\limits_{u \in \mathcal{N}(v)}M^{(k)}\left(\mathbf h_v^{(k-1)},\mathbf h_u^{(k-1)},\mathbf x_{v}^g, \mathbf x_{u}^g, \mathcal E_{v,u}\right)\textbf{(Node)},\\
     \sum\limits_{u \in \mathcal{N}(v)}M^{(k)}\left(\mathbf h_v^{(k-1)},\mathbf h_u^{(k-1)},\mathbf S_{v,u}^g, \mathcal E_{v,u}\right)\textbf{(Edge)}.
\end{cases}
\end{aligned}
\end{equation}

GSN is a promising start for subgraph-enhanced MPNNs.
However, they have one fatal drawback in that searching for and testing subgraphs for isomorphism is computationally prohibitive. 
To avoid this high computational bottleneck, GNN-AK \cite{zhao2022from} samples subgraphs and swift encodes them into node embeddings. 
Specifically, GNN-AK extracts the neighborhoods of each node as subgraphs (i.e., the neighborhood of node $v$ is a subgraph $g_v$), and applies a base MPNN to each neighborhood subgraph to obtain the final node embeddings, i.e.:
\begin{equation}
\begin{aligned}
    \mathbf x_v^g =[Emb\left(v|g_v\right) \mid \sum_{u\in \mathcal V \land  u \neq v} Emb\left(u|g_v\right) \mid \sum_{u\in \mathcal V \land  u \neq v} Emb\left(v|g_u\right)], 
\end{aligned}
\end{equation}
where $Emb\left(v|g_v\right)$ is the embedding of node $v$ produced by running the base MPNN on subgraph $g_v$, $Emb\left(v|g_u\right) == 0$ if subgraph $g_u$ does not contain node $v$ (i.e., $v \not\subset \mathcal V_{g_u}$), and $\mathbf x_v^g$ is the subgraph feature of node $v$ for MPNN's aggregation.

Analogously, Nested Graph Neural Networks (NGNN) \cite{zhang2021nested} extracts nodes (i.e., $\mathcal N(v) \cup v$) and edges (i.e., $\mathcal E_{v_1,v_2} \in \mathcal E \And v_1,v_2 \in \mathcal N(v) \cup v$) in the 1-hop neighborhood of node $v$, as a neighborhood subgraph $g_v$, to be encoded by a GNN. 
The subgraph $g_v$ is then encoded as the embedding $h_{g_v}$, which denotes the subgraph feature of node $v$.

One thing common to all the above methods is that they dilute or replace the node features. But such feature properties are essential for graph-level learning. 
Thus, GraphSNN \cite{wijesinghe2021new} incorporates the idea of encoding the subgraph features into the edge's weight for aggregation without changing the node features. 
This approach defines the formula for calculating the degree of isomorphism between two subgraphs. 
The weight of $\mathcal E_{v,u}$ is equal to the degree of isomorphism between two specific subgraphs, where one of the subgraphs is the node $v$'s neighborhood subgraph, and the other subgraph is the overlap between the neighborhood subgraphs of nodes $u$ and $v$.
By normalizing the computed weights at the end, GraphSNN builds a subgraph-guided attention mechanism partaking in the MPNN's aggregation.

\subsubsection{Graphlet} In addition to empowering MPNNs through subgraph information, researchers have directly used the embeddings of subgraphs to form a graph-level representation.
SUGAR \cite{sun2021sugar}, for example, uses GNNs to embed discriminative subgraphs selected through reinforcement learning.
A readout function over all learned subgraph embeddings is then used to build a graph-level representation for classification, which can be used for classification. 
Correspondingly, the graph classification results are back-propagated to train the GNNs that embed selected subgraphs.
Similarly, Subgraph Neural Networks (SubGNN) \cite{AlsentzerFLZ20subgraph} views the subgraphs of a graph as instances with independent labels. 
For each instance, SubGNN samples a few finer local structures, and forms embeddings through the GNN.
A representation of each instance is generated by aggregating all the embeddings of the sampled local structures.
Another approach, Equal Subgraph Aggregation Network (ESAN) \cite{bevilacqua2021equivariant}, enhances this branch by applying two GNNs, one for learning individual embeddings for sampled subgraphs and the other for learning message passing among them. Finally, a universal set encoder \cite{qi2017pointnet} compresses all the subgraph embeddings into one graph-level representation.

\subsection{Graph Kernel-Based GL-GNNs}\label{kernel-based GL-GNNs} 

Like the revival of the subgraph idea in the deep learning field, graph kernels that incorporate deep learning techniques have also attracted attention.
Similar to subgraph-based GL-GNNs, there are generally two branches of graph kernel-based GL-GNNs. 
As Fig. \ref{fmpnn} (C) shows, one branch replaces the 1-hop neighbor aggregation and vertex update functions in MPNNs with a kernel function.
This group is, called the kernel-enhanced MPNNs.
In the other branch, differentiable and parameterizable kernels are designed to plug kernels into the neural networks so as to form learnable and fast deep graph kernels.

\subsubsection{Kernel-enhanced MPNNs} This type of method often uses a graph kernel to update the node embeddings, which in turn are used to recalculate the graph kernel. 
Kernel-enhanced MPNNs break the local update of the MPNN (i.e., the node features are aggregated via adjacent neighbors) so as to capture more structure information.

For example, Graph Convolutional Kernel Networks (GCKN) \cite{chen2020convolutional} and Graph Structured Kernel Networks (GSKN) \cite{long2021theoretically} employ walk-based kernels to iteratively generate node embeddings. 
Specifically, these methods generate $q$ walking sequences starting from the target node, where each sequence records all node embeddings in the walk.
As an example, in the $k$-th iteration, the one-step walk sequence $P_{i}$ from node $v$ to $u$ would be represented as $R(P_{i}) = [\mathbf h_v^{(k-1)},\mathbf h_u^{(k-1)}]^\top$.
By building a kernel function $K\left(R(P_{i}),R(P_{j})\right)$ (e.g., a random walk kernel) as the similarity measurement for any two walking sequences $P_{i}$ and $P_{j}$, GCKN and GSKN aggregate the kernel values as the updated node embeddings, that is:
\begin{equation}
    h_v^{(k)} = \sum_{1 \leq i \leq q} [K\left(R(P_{1}),R(P_{i})\right), \cdots, K\left(R(P_{q}),R(P_{i})\right)]^\top.
\end{equation}
To follow up, the node embeddings updated by the graph kernel are used to obtain the kernel value in the next iteration.
Du \textit{et al.} \cite{du2019graph} combined a Neural tangent kernel (NTK) \cite{jacot2018neural} with an MPNN, summarizing the advantages of this category of approach. 
Overall, the technique gives better theoretical explanations, brought about by the graph kernel, and the convex-optimized tasks are easy to train.
Thus, kernel-enhanced MPNNs use a kernel function to replace the aggregation and vertex update functions in MPNNs. 
The walk-based kernels do particularly well at capturing local structures to encode into the node embeddings.

\subsubsection{Deep Graph Kernel} Traditional graph kernels are limited by the theoretical computational bottleneck, thus, researchers search for an optimal solution for comparing two graphs by neural networks.
Recently, Lei \textit{et al.} \cite{lei2017deriving} discussed deep graph kernels as parameterized learnable graph kernels for deriving neural operations.
These deep graph kernels can be optimized for specific tasks with fast computation speeds and good interpretability.

Deep Divergence Graph Kernels (DDGK) \cite{al2019ddgk} takes $M$ base graphs $\{\mathcal G_1, \mathcal G_2, \cdots, \mathcal G_M\}$ to represent a target graph $\mathcal G_t$ as $M$-dimensional vectors $\mathbf h_{\mathcal G_t} = [K_D \left(\mathcal G_1, \mathcal G_t\right), \cdots, K_D \left(\mathcal G_M, \mathcal G_t\right)]^\top$, where $K_D \left(\mathcal G_m, \mathcal G_t\right)$ is a trainable kernel for measuring the distance between $\mathcal G_m$ and $\mathcal G_t$. 
First, DDGK uses each base graph $\{\mathcal G_1, \mathcal G_2, \cdots, \mathcal G_M\}$ to train an encoder $\{\mathcal Z_1, \mathcal Z_2, \cdots, \mathcal Z_M\}$.
The encoder $\mathcal Z_m$ takes the one-hot encoding of nodes (e.g., the first node's encoding is $[1, 0, 0, \cdots]^\top$) in $\mathcal G_m$ as the input and tries to predict their neighbors (e.g., if a node only links to the second and third nodes, the correct output should be $[0, 1, 1, 0, \cdots]^\top$).
Then, the trained encoder $\mathcal Z_m$ is used for predicting the node's neighbors in $\mathcal G_t$, as the divergence score $K_D \left(\mathcal G_m, \mathcal G_t\right)$ between two graphs.
That is:
\begin{equation}
    K_D \left(\mathcal G_m, \mathcal G_t\right) = \sum\limits_{v_i,v_j \in \mathcal V_t, \mathcal E_{i,j} \in \mathcal E_t} -\log\left(v_j | v_i, \mathcal Z_m\right). 
\end{equation}

Random Walk graph Neural Networks (RWNN) \cite{NEURIPS2020_ba95d78a} also derives a trainable random walk kernel $K_{RW}\left(\cdot, \cdot\right)$ through a series of learnable graph patterns $\{\mathcal G_1, \mathcal G_2, \cdots, \mathcal G_M\}$.
A learnable graph $\mathcal G_m$ has a fixed node set $\mathcal V_{m}$ but a changeable edge set $\mathcal E_{m}$.
RWNN produces graph-level embeddings $\mathbf h_{\mathcal G_t} = [K_{RW} \left(\mathcal G_1, \mathcal G_t\right), \cdots, K_{RW} \left(\mathcal G_M, \mathcal G_t\right)]^\top$ of the target graph $\mathcal G_t$ for graph classification tasks.
Correspondingly, the classification results are backpropagated to change the adjacency matrix of learnable graph patterns.
That is, RWNN uses the prediction results to train the input of the kernel function (i.e., graph patterns) so that the kernel values can be learned according to the downstream task.

\subsection{Spectral-Based GL-GNNs} \label{spectralgnn}

Spectral-based GL-GNNs were started earlier by Bruna \textit{et al.} \cite{Spectral2014Bruna}, which designed graph convolutions via the spectral graph theory \cite{chung1997spectral}.
Recently, Balcilar \textit{et al.} \cite{balcilar2020analyzing} described spectral and spatial graph convolution in a unified way and performed spectral analysis on convolution kernels. 
The analysis results demonstrate that a vast majority of MPNNs are low-pass filters in which only smooth graph signals are retained. Graph signals with a low-frequency profile are useful for node-level tasks on assortative networks where nodes have similar features to their neighborhoods \cite{bo2021beyond}. However, with graph-level tasks, graph signals beyond the low frequency may be critical since they can highlight the differences between different graphs \cite{balcilar2020bridging}, and, although MPNNs have been widely used, they overlook the signal frequency of graph data.

In terms of a feature $\mathbf x\in \mathbb R^n$ (a column vector of $\mathbf X\in \mathbb R^{n\times f}$) as a graph signal on a graph with $n$ nodes, spectral graph convolution performs graph signal filtering after transforming the graph signals $\mathbf x$ in spatial space into the frequency domain. According to spectral graph theory \cite{chung1997spectral}, the frequency domain generally takes the eigenvectors of the graph Laplacian $\mathbf L=\mathbf D-\mathbf A$ where $\mathbf D$ is the degree matrix (or the normalized version $\mathbf L = \mathbf I- \mathbf D^{-\frac{1}{2}}\mathbf A \mathbf D^{-\frac{1}{2}}$) of a set of space bases.
Note, though, that other bases can also be used, such as graph wavelet bases \cite{xu2019graph,hammond2011wavelets}. Specifically, $\{\lambda_1, ..., \lambda_n\}$ where $0 \leq \lambda_1 \leq ... \leq \lambda_n \leq 2$, and $\mathbf U=(\mathbf u_1, ..., \mathbf u_n)$ are the $n$ eigenvalues and $n$ orthogonal eigenvectors of $\mathbf L$, respectively. $\lambda_i$ represents the smoothness degree of $\mathbf u_i$ about $\mathbf L$. Based on the graph Fourier transformation $\hat{\mathbf x} =\mathbf U^\mathrm{T}\mathbf x$, the graph signal $\mathbf x$ is mapped to the frequency domain. And $\mathbf x=\mathbf U\hat{\mathbf x}$ is the graph Fourier inverse transformation that can restore the graph signal in spectral domain to the spatial domain. The polynomial filter is adopted by most of spectral graph convolution methods, for example, ChebNet \cite{defferrard2016convolutional} defines the spectral graph convolution as $\mathbf U \text{diag}(\Phi(\mathbf \Lambda))\mathbf U^\mathrm{T}\mathbf x$, where $\Lambda=\text{diag}(\{\lambda_i\}_{i=1}^{i=n})$, $\Phi(\mathbf \Lambda)=\sum_{k=0}^{K}\theta_k\mathbf \Lambda^{k}$ is the polynomial filtering function, $K$ are the hyper-parameters that realize the localized spectral graph convolution, and $\theta_k$ is the polynomial coefficient. 

Spectral graph convolution can be task-agnostic when graph signals with any frequency profiles are filtered. 
Conversely, they can also be task-specific \textemdash for example, a band-pass filter can highlight graph signals that are strongly relate to downstream tasks \cite{balcilar2020analyzing}. 
However, only a few practitioners have designed graph-level neural networks from the perspective of spectral graph convolution \cite{balcilar2021breaking, bianchi2021graph}.
The main problem with applying spectral convolution in graph-level tasks is the transferability of the spectral filter coefficients from the training graph set to the unseen graphs. The spectral filters depend on the graph Laplacian decomposition, but different graph structures have different graph Laplacian decomposition results. 
Most recently, Levie \textit{et al.} \cite{levie2021transferability} theoretically proved the transferability of spectral filters on multigraphs. 
Balcilar \textit{et al.} \cite{balcilar2021breaking} proposed a custom filter function that could output frequency components from low to high to better distinguish between graphs. 
Due to the limitation of polynomial filters in modeling sharp changes in the frequency response, Bianchi \textit{et al.} \cite{bianchi2021graph} employed an auto-regressive moving average (ARMA) filter to perform graph-level tasks. The ARMA filter is more robust to the changes or perturbations on graph structures as it does not depend on the eigen-decomposition of the graph Laplacian explicitly. 
In addition, Zheng \textit{et al.} \cite{zheng2021framelets} proposed a graph convolution based on graph Framelet transforms instead of graph Fourier transform with a shrinkage activation to decompose graphs into both low-pass and high-pass frequencies. However, there is no theoretical proof of the transferability of framelet decomposition.

%% file: 7-GLNN-Pooling.tex
\section{Graph Pooling}\label{Pooling}

Generally, deep graph-level learning methods encode graphs based on node representations. Graph pooling is a technique that integrates node embeddings into a graph embedding. In this section, we introduce two mainstream types of graph pooling techniques, i.e., global and hierarchical graph pooling (see Section \ref{global} and \ref{hierarchical}). We summarize all discussed pooling approaches in Table \ref{table_graphpooling} in Appendix \ref{ap_Pooling}. Moreover, we discuss the effectivity of graph pooling (see Section \ref{effectiveness}).

\subsection{Global Graph Pooling}\label{global}

There are four different types of global graph pooling \textemdash numeric operation, attention-based, CNN-based, and global top-$K$ \textemdash all of which aggregate all node embeddings at once to build a graph-level representation.

\subsubsection{Numeric Operation} Adopting a simple numeric operation for all node embeddings is a common graph pooling method \cite{xu2018powerful,duvenaud2015convolutional}, since it is easy to use and obeys the permutation invariant. 
An illustration of a type of numeric operation (i.e., a summation) for all node embeddings is shown in Fig. \ref{fglobalpool} (A).
It is common to see practitioners aggregating node embeddings via summation, maximization, minimization, mean, and concatenation functions. For example:
\begin{equation}
    \mathbf h_{\mathcal G} = \sum_{v \in \mathcal V} \mathbf h_v \bigggl / \mathop{\text{max/min}}\limits_{v \in \mathcal V} \left(\mathbf h_v\right) \bigggl / \frac{1}{|\mathcal V|}\sum_{v \in \mathcal V} \mathbf h_v \bigggl / [\mathbf h_{v_1}|...|\mathbf h_{v_{|\mathcal V|}}].
\end{equation}

Duvenaud \textit{et al.} \cite{duvenaud2015convolutional} empirically proved that, in graph-level learning, summation has no weaker an outcome than a hash function. 
Similarly, GIN \cite{xu2018powerful} shows us that the injective relabeling function in the WL algorithm can be replaced with a simple numeric operation. 
Further, GIN also allows us to analyze the efficacy of different functions: summation, maximization, and mean functions. 
Summation comprehensively summarizes the full features and structure of a graph.
Maximization emphasizes significant node embeddings, and mean learns the distribution of labels. 
Inspired by GIN, Principal Neighbourhood Aggregation (PNA) \cite{NEURIPS2020_99cad265} employs all three of these functions to pool the node embeddings, while TextING \cite{zhang2020every} includes both mean and maximization pooling to capture the label distribution and strengthen the keyword features. 
A few variants of graph pooling have also been developed.
For example, Deep Tensor Neural Network (DTNN) \cite{schutt2017quantum} applies a neural layer that processes the node embeddings before the summation function and second-order pooling (SOPOOL) \cite{wang2020second} is executed as $\mathbf h_{\mathcal G} = [\mathbf h^T_{v_1}\mathbf h_{v_1}|...|\mathbf h^T_{v_{|\mathcal V|}}\mathbf h_{v_{|\mathcal V|}}]$.

\subsubsection{Attention-based} The contributions of node embeddings to graph-level representations may not be equal, as some of them contain may more important information than others.
Hence, some researchers have tried using an attention mechanism to aggregate the node embeddings based on their particular contribution, as outlined in Fig. \ref{fglobalpool} (C).
Li \textit{et al.} \cite{li2016gated} and Duvenaud \textit{et al.} \cite{duvenaud2015convolutional}, for example, both employ a softmax function as an attention-based global pooling for aggregation. 
This can be written as:
\begin{equation}
\mathbf h_{\mathcal G} = \sum_{v,k} \text{softmax} \left(w_v^k, \mathbf h_v^k\right), 
\end{equation}
where $w_v^k$ is a trainable weight for the embedding $h_v^k$ of node $v$ in iteration $k$.
Note that $w_v^k$ will be large if $h_v^k$ is important to the downstream task.
Set2Set \cite{set2set} is a more complicated attention-based graph pooling model. It learns the attention coefficients of all node embeddings from an ordered sequence generated by LSTM. Although Set2Set handles sequential node embeddings, the order of nodes is determined by an LSTM model without affecting permutation invariance.

\subsubsection{CNN-based} PATCHY-SAN \cite{niepert2016learning} and KCNN \cite{nikolentzos2018kernel} are based on the idea of ordering vertices and applying a 1-D convolutional layer to pool the ordered vertices features. 
These two models are permutation invariant because they order vertices according to certain rules regardless of the input order.

\subsubsection{Global Top-$K$} 
\begin{figure*}[htbp!]
    \centering
    \includegraphics[width=1\textwidth]{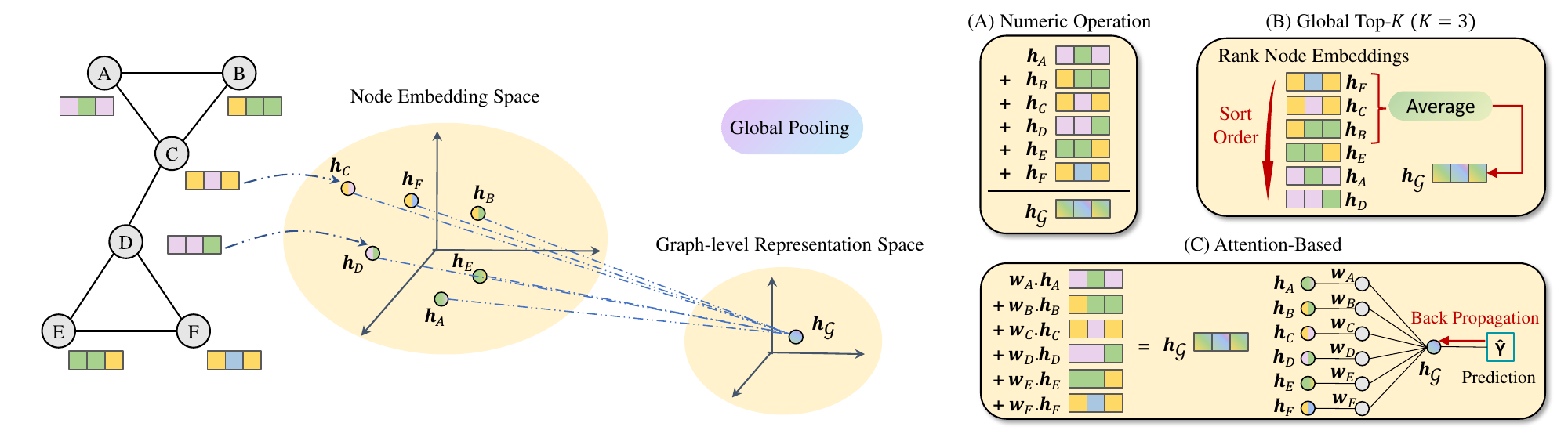}
    \caption{Toy examples of Global Pooling methods.}
    \label{fglobalpool}
\end{figure*}

Global top-$K$ graph pooling sorts all nodes and selects the first $K$ node embeddings for aggregation, as shown in Fig. \ref{fglobalpool} (B).
In this way, the pooling layer only preserves $K$ significant vertices and drops out others.
SortPool \cite{zhang2018end} employs graph convolution operations to project each node into a one-dimensional vector as the ranking score for selecting the $K$ vertices with the highest scores.
Subsequently, a GL-GNN is used to produce the node embeddings of the selected $K$ nodes, which come together to form the graph-level representation.
Graph Self-Adaptive Pooling (GSAPool) \cite{zhang2020structure} is another global top-$K$ graph pooling model that ranks nodes based on the summing of feature and structure scores. 
The node structure scores are 1-dimensional vectors projected by the graph convolution operations as same as SortPool, while the feature scores are learned by feeding the node features into an MLP.

\subsection{Hierarchical Graph Pooling}\label{hierarchical}

\begin{figure*}[htbp!]
    \centering
    \includegraphics[width=1\textwidth]{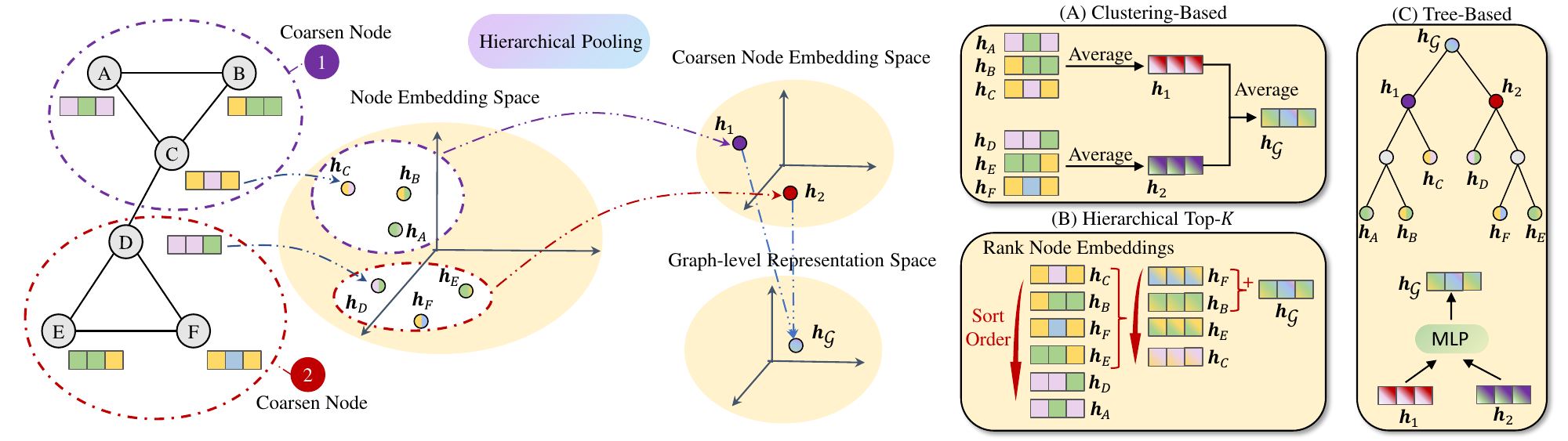}
    \caption{Toy examples of Hierarchical Pooling methods.}
    \label{fhierpool}
\end{figure*}

Global graph pooling ignores the hierarchical structures in graphs.
The evolution of a graph is to collect nodes into hierarchical structures (e.g., communities), then to form the graph.
Hence, researchers tend to capture hierarchical information through an aggregation process that has multiple parses, which coarsens the graph each time.
We have divided hierarchical graph pooling techniques into three branches: clustering-based, hierarchical top-$K$, and tree-based.

\subsubsection{Clustering-based} Clustering methods were originally designed to capture the hidden hierarchical structures in graphs, but these techniques can be incorporated into the pooling process.
Fig. \ref{fhierpool} (A) demonstrates clustering-based graph pooling, which has been the focus of many studies.
For instance, Henaff \textit{et al.} \cite{henaff2015deep} implemented multi-resolution spectral clustering \cite{von2007tutorial} which assigns each node to a matched cluster. 
Subsequently, the clusters in the input graph are treated as new nodes of the new coarsened graph.
The embedding of the new node is obtained by averaging all node embeddings in the cluster.
This coarsening process is iterative and operates until only one or very few vertices remain in the most recent coarsened graph.
Similarly, Bruna \textit{et al.} \cite{ICLR2014BRUNA} adopted hierarchical agglomerative clustering \cite{hastie2009elements} to coarsen graphs, while StructPool \cite{yuan2020structpool} employs conditional random fields \cite{lafferty2001conditional} to cluster each node by considering the assignments of other vertices.

However, clustering-based graph pooling cannot optimize the clustering process for downstream tasks given just any old off-the-shelf clustering method.
Rather, the clustering method must be designed to consider downstream tasks.
For example, Graph Multiset Transformer (GMT) \cite{baek2020accurate} uses a multi-head self-attention mechanism to cluster nodes into different sets according to the final task and a graph-level representation is therefore derived through these sets.
MinCutPool \cite{bianchi2020spectral} assigns each node to a cluster via an MLP, which is optimized by two goals: first that the clusters are similar in size, and, second, that the clusters' embeddings are separable.
Finally, the graph-level representation is obtained by pooling the substructure-level embeddings.
EigenPool \cite{ma2019graph} involves a spectral clustering method that coarsens graphs and pools node embeddings into cluster-level embeddings by converting spectral-domain signals.
These clustering-based algorithms assume that each node belongs to a certain cluster, yet DiffPool \cite{ying2018hierarchical} assigns each node to multiple clusters through a trainable soft assignment matrix $\mathbf S^{(k)} \in \mathbb R^{n^{(k)} \times n^{(k+1)}}$, where $n^{(k)}$ is the number of vertices in the input graph at the $k$-th layer, and $n^{(k+1)}$ represents the cluster's number in the input graph or the node's number in the coarsened graph.
To be specific, at the $k$-th layer, each row of $\mathbf S^{(k)}$ corresponds to a node in the input graph, and each column of $\mathbf S^{(k)}$ corresponds to a new node in the coarsened graph (i.e., a cluster in the input graph). 
The assignment matrix $\mathbf S^{(k)}$ is trained by a graph convolutional layer, which is defined as:
\begin{equation}
    \mathbf S^{(k)} = \text{softmax} \left(\text{Conv}^{(k)} \left(\mathbf A^{(k)},\mathbf H^{(k)},\mathbf W^{(k)}\right)  \right), 
\end{equation}
where $\mathbf A^{(k)} \in \mathbb R^{n^{(k)} \times n^{(k)}}$ and $\mathbf H^{(k)} \in \mathbb R^{n^{(k)} \times f}$ are the adjacent matrix and node embedding matrix of the input graph at the $k$-th layer, respectively. $\mathbf W^{(k)} \in \mathbb R^{f \times n^{(k+1)}} $ is the trainable weight matrix, and $\text{softmax}\left(\cdot\right)$ function is applied to each row.

\subsubsection{Hierarchical Top-$k$} The high complexity of the clustering process exacerbates the computational cost burden of cluster-based hierarchical graph pooling.
For example, the DiffPool \cite{ying2018hierarchical} is extremely costly in terms of time and memory because the assignment matrices need to be trained. 
So, to speed up the process of hierarchical graph pooling, researchers have looked to replace the clustering process with a scheme that coarsens the graph according to the top-$K$ idea, as shown in Fig. \ref{fhierpool} (B).
Graph U-nets \cite{gao2019graph}, for example, projects each node feature into a 1-dimensional vector $\mathbf Y$, as the rank score. Subsequently, the $K$ nodes with the highest score are selected to form the new coarsened graph, which is defined as:
\begin{equation}
\begin{aligned}
    \mathbf Y = \frac{\mathbf Z^{(l)}\mathbf P^{(l)}}{||\mathbf P^{(l)}||}, 
    \quad 
    \text{idx} = \text{Top}\ K\left(\mathbf Y\right), 
    \quad
    \mathbf Z^{(l+1)} = \left(\mathbf Z^{(l)} \odot \left(\text{sigmoid}\left(\mathbf Y_{\text{idx}} \right) \mathbf 1^\top_Z\right) \right),  
    \quad
    \mathbf A^{(l+1)} = \mathbf A^{(l)}_{\text{idx},\text{idx}},
\end{aligned}
\end{equation}
where $Z^{(l)}$ is the input node features at the layer $l$, $\mathbf P^{(l)}$ is a learnable projection matrix, $\text{Top}\ K\left(\cdot \right)$ is a function that returns the index of the top-$K$ nodes, and all the elements are 1 in the vector $\mathbf 1_Z$ which has the same dimension as the node feature.
Cangea \textit{et al.} \cite{cangea2018towards} employed the Graph U-nets to coarsen graphs and concatenated the mean and maximum values of node embeddings on the coarsened graphs as graph-level representations.
Further, SAGPool \cite{lee2019self} chooses the top-$K$ nodes to generate the coarsened graph by adopting a graph convolution operation to project node features as scores.

All these methods generate a coarsened graph by preserving the top-$K$ nodes. 
However, Ranjan \textit{et al.} \cite{ranjan2020asap} presented the novel idea of ranking the clusters and preserving on the top-$K$ of them.
The clusters were ranked by employing a self-attention algorithm called Master2Token \cite{shen2018disan} that scores each cluster based on the node embeddings within it.

\subsubsection{Tree-based} Tree-based hierarchical graph pooling implements the coarsening process via an encoding tree, where the input graph is coarsened layer by layer to the ultimate node from the leaf layer to the root layer, as shown in Fig. \ref{fhierpool} (C).
ChebNet \cite{defferrard2016convolutional} and MoNet \cite{monti2017geometric} use the Graclus \cite{dhillon2007weighted} algorithm to pair nodes in the graph based on the graph spectrum and merge the pair-wise nodes as a new node in the coarsened graph.
That is to say, these two methods build a balanced binary tree to coarsen the graph, and each father node on the tree is obtained by coarsening its two child nodes.
Wu \textit{et al.} \cite{wu2021structural} uses a structure encoding tree \cite{li2016structural} for tree-based hierarchical graph pooling.
Structural coding trees compress the hierarchy of a graph into a tree. Here, the leaves are the nodes, the root represents the whole graph, and the other non-leaf nodes are the hierarchical structures (e.g., the communities).
An MLP merges the features of the child nodes in the structure encoding tree, to generate an embedding of the father node.
The result is an embedding of the root node, which serves as a graph-level representation.
Moreover, Wu \textit{et al.} \cite{wu2022structural} empirically verified that the hierarchical tree pooling guided by structure entropy can preserve higher-quality structural information than U-Nets and MinCutPool.
Alternatively, EdgePool \cite{diehl2019edge} scores edges based on the features of the nodes the edges link, eliminating the highest-ranked edge by merging its two end nodes.
The features of the newly generated node, which maintains all the neighbors of the original two nodes, are obtained by summing the features of the two merged nodes.
EdgePool falls into the category of being a tree-based hierarchical graph pooling method because it merges two child nodes in a tree into a father node.

%% file: 8-Datasets_and_Evaluations.tex
\section{Benchmarks}

\begin{table*}[]
\centering
\caption{Summary of Selected Benchmark Datasets}\label{table_data}
\begin{threeparttable}
\resizebox{\linewidth}{!}{
\begin{tabular}{llllllllcl}
\hline
Category                                                                    & Dataset          & Size   & \#Graphs & \begin{tabular}[c]{@{}l@{}}Average\\ \#Nodes\end{tabular} & \begin{tabular}[c]{@{}l@{}}Average\\ \#Edges\end{tabular} & \begin{tabular}[c]{@{}l@{}}Node\\ Attr.\end{tabular} & \begin{tabular}[c]{@{}l@{}}Edge\\ Attr.\end{tabular} & \#Classes & Source                                                                           \\ \hline
\multirow{6}{*}{Biology}                                                    & ENZYMES          & Small  & 600      & 32.6                                                      & 62.1                                                      & \checkmark                                            & -                                                    & 6         & \cite{borgwardt2005protein,morris2020tudataset}                         \\
                                                                            & PROTEINS         & Small  & 1113     & 39.1                                                      & 72.8                                                      & \checkmark                                            & -                                                    & 2         & \cite{borgwardt2005protein,morris2020tudataset}                         \\
                                                                            & D\&D             & Small  & 1178     & 284.3                                                     & 715.7                                                     & \checkmark                                            & -                                                    & 2         & \cite{dobson2003distinguishing,morris2020tudataset}                     \\
                                                                            & BACE             & Small  & 1513     & 34.1                                                      & 36.9                                                      & \checkmark                                            & \checkmark                                            & 2         & \cite{wu2018moleculenet,subramanian2016computational}                   \\
                                                                            & MUV              & Medium & 93087    & 24.2                                                      & 26.3                                                      & \checkmark                                            & \checkmark                                            & 2         & \cite{wu2018moleculenet,rohrer2009maximum}                              \\
                                                                            & ppa              & Medium & 158100   & 243.4                                                     & 2266.1                                                    & -                                                    & \checkmark                                            & 37        & \cite{zitnik2019evolution,hu2020open}                                   \\ \hline
\multirow{14}{*}{Chemistry}                                                 & MUTAG            & Small  & 188      & 17.9                                                      & 19.8                                                      & \checkmark                                            & \checkmark                                            & 2         & \cite{10.5555/3042573.3042614,morris2020tudataset} \\
                                                                            & SIDER            & Small  & 1427     & 33.6                                                      & 35.4                                                      & \checkmark                                            & \checkmark                                            & 2         & \cite{wu2018moleculenet,altae2017low}                                   \\
                                                                            & ClinTox          & Small  & 1477     & 26.2                                                      & 27.9                                                      & \checkmark                                            & \checkmark                                            & 2         & \cite{wu2018moleculenet,gayvert2016data}          \\
                                                                            & BBBP             & Small  & 2039     & 24.1                                                      & 26.0                                                      & \checkmark                                            & \checkmark                                            & 2         & \cite{wu2018moleculenet,martins2012bayesian}                            \\
                                                                            & Tox21            & Small  & 7831     & 18.6                                                      & 19.3                                                      & \checkmark                                            & \checkmark                                            & 2         & \cite{wu2018moleculenet,challengetox21}                                 \\
                                                                            & ToxCast          & Small  & 8576     & 18.8                                                      & 19.3                                                      & \checkmark                                            & \checkmark                                            & 2         & \cite{wu2018moleculenet,richard2016toxcast}                             \\
                                                                            & MolHIV           & Small  & 41127    & 25.5                                                      & 27.5                                                      & \checkmark                                            & \checkmark                                            & 2         & \cite{wu2018moleculenet,hu2020open}                                     \\
                                                                            & MolPCBA          & Medium & 437929   & 26.0                                                      & 28.1                                                      & \checkmark                                            & \checkmark                                            & 2         & \cite{wu2018moleculenet,hu2020open}                                     \\
                                                                            & FreeSolv         & Small  & 642      & 8.7                                                       & 8.4                                                       & \checkmark                                            & \checkmark                                            & -         & \cite{wu2018moleculenet,mobley2014freesolv}                             \\
                                                                            & ESOL             & Small  & 1128     & 13.3                                                      & 13.7                                                      & \checkmark                                            & \checkmark                                            & -         & \cite{wu2018moleculenet,delaney2004esol}                                \\
                                                                            & Lipophilicity    & Small  & 4200     & 27.0                                                      & 29.5                                                      & \checkmark                                            & \checkmark                                            & -         & \cite{wu2018moleculenet,wenlock2015experimental}                        \\
                                                                            & AQSOL            & Small  & 9823     & 17.6                                                      & 35.8                                                      & \checkmark                                            & \checkmark                                            & -         & \cite{dwivedi2020benchmarking,sorkun2019aqsoldb}                        \\
                                                                            & ZINC             & Small  & 12000    & 23.2                                                      & 49.8                                                      & \checkmark                                            & \checkmark                                            & -         & \cite{dwivedi2020benchmarking,irwin2012zinc}                            \\
                                                                            & QM9              & Medium & 129433   & 18.0                                                      & 18.6                                                      & \checkmark                                            & \checkmark                                            & -         & \cite{wu2018moleculenet,morris2020tudataset}    \\ \hline
\multirow{7}{*}{\begin{tabular}[c]{@{}l@{}}Social \\ Networks\end{tabular}} & IMDB-BINARY      & Small  & 1000     & 19.8                                                      & 96.5                                                      & -                                                    & -                                                    & 2         & \cite{Yanardag2015kdd,morris2020tudataset}                              \\
                                                                            & IMDB-MULTI       & Small  & 1500     & 13.0                                                      & 65.9                                                      & -                                                    & -                                                    & 3         & \cite{Yanardag2015kdd,morris2020tudataset}                              \\
                                                                            & DBLP\_v1         & Small  & 19456    & 10.5                                                      & 19.7                                                      & \checkmark                                            & \checkmark                                            & 2         & \cite{morris2020tudataset}                                              \\
                                                                            & COLLAB           & Medium & 5000     & 74.5                                                      & 2457.8                                                    & -                                                    & -                                                    & 3         & \cite{Yanardag2015kdd,morris2020tudataset}                              \\
                                                                            & REDDIT-BINARY    & Small  & 2000     & 429.6                                                     & 497.8                                                     & -                                                    & -                                                    & 2         & \cite{Yanardag2015kdd,morris2020tudataset}                              \\
                                                                            & REDDIT-MULTI-5K  & Medium & 4999     & 508.5                                                     & 594.9                                                     & -                                                    & -                                                    & 5         & \cite{Yanardag2015kdd,morris2020tudataset}                              \\
                                                                            & REDDIT-MULTI-12K & Medium & 11929    & 11.0                                                      & 391.4                                                     & -                                                    & -                                                    & 11        & \cite{Yanardag2015kdd,morris2020tudataset}                              \\ \hline
\multirow{4}{*}{\begin{tabular}[c]{@{}l@{}}Computer\\ Science\end{tabular}} & CIFAR10          & Medium & 60000    & 117.63                                                    & 941.1                                                     & \checkmark                                            & -                                                    & 10        & \cite{dwivedi2020benchmarking,krizhevsky2009learning}                   \\
                                                                            & MNIST            & Medium & 70000    & 70.57                                                     & 564.53                                                    & \checkmark                                            & -                                                    & 10        & \cite{dwivedi2020benchmarking,lecun1998mnist}                           \\
                                                                            & code2            & Medium & 452741   & 125.2                                                     & 124.2                                                     & \checkmark                                            & \checkmark                                            & -         & \cite{husain2019codesearchnet,hu2020open}                               \\
                                                                            & MALNET           & Large  & 1262024  & 15378                                                     & 35167                                                     & -                                                    & -                                                    & 696       & \cite{freitas2021a}                                                     \\ \hline
\end{tabular}}
\vspace{+1mm}
\begin{tablenotes}
 \small
 \item[*] The category of computer science includes computer vision, cybersecurity, and program coding datasets.
 \item[*] Node Attr. and Edge Attr. indicates the labels or features of nodes and edges, respectively.
 \item[*] The size of datasets follows the setting of OGB \cite{hu2020open}, medium datasets have more than 1 million nodes or more than 10 million edges, and large datasets own over 100 million nodes or 1 billion edges.
\end{tablenotes}
\end{threeparttable}
\end{table*}


\subsection{Datasets}
Table \ref{table_data} summarizes a selection of benchmark graph-level datasets, including TUDateset \cite{morris2020tudataset}, Open Graph Benchmark (OGB) \cite{hu2020open}, MOLECULENET \cite{wu2018moleculenet}, MALNET \cite{freitas2021a}, and others \cite{dwivedi2020benchmarking}.
The graph datasets collected by the group at TUDateset \cite{morris2020tudataset} have been widely used to evaluate graph-level learning approaches. 
These graph datasets consist of molecules, proteins, images, social networks, synthetic graphs, and data from many other domains. 
However, despite their wide use, they have attracted criticism from some practitioners. 
For example, Ivanov \textit{et al}. \cite{ivanov2019understanding} contends that the sets suffer from isomorphism bias, i.e., they contain isomorphic graphs with different labels, which may hinder model training \textemdash a claim based on the analysis of 54 widely-used graph datasets.
They also note that some of the datasets are too small to train a data-hungry deep learning model. 
For example, Dwivedi \textit{et al.} \cite{dwivedi2020benchmarking} presented that most GL-GNNs have a close performance to others in the small dataset. 
Further, some topology-agnostic baselines yield a performance that is competitive to GL-GNNs. 

Developing practical and large-scale benchmark datasets has become an important issue for the graph-level learning community.
To this end, Wu et al. \cite{wu2018moleculenet} proposed a benchmark named MOLECULENET that contains a set of large-scale graph datasets of molecules.
The dataset is designed to be used for graph regression and classification tasks. Dwivedi \textit{et al.} \cite{dwivedi2020benchmarking} transformed images into graphs for classification, in which a group of pixels is clustered as a node. Based on real-world cybersecurity scenarios, Freitas \textit{et al.} \cite{freitas2021a} proposed a large-scale graph dataset of over 1.2 million graphs with imbalanced labels. 
Furthermore, OGB \cite{hu2020open} has published application-oriented large-scale graph datasets of molecules, proteins, and source code cooperation networks.

\subsection{Evaluations}

The development of graph-level learning has been impeded by unfair evaluations. 
For example, Ruffinelli \textit{et al.} \cite{ruffinelli2019you} argue that some graph-level learning models only produce state-of-the-art performance because of tricks with the model's training, not because of the novel ideas proposed in the articles. However, there is no consensus on which evaluation to use with the most widely used graph datasets, such as TUDatasets, nor is there even a universally-accepted data split \cite{Errica2020A}. 
Hence, to evaluate the graph-level learning models in a unified and fair way, some researchers have attempted to establish a standard model evaluation protocol. 
For example, Dwivedi \textit{et al.} \cite{dwivedi2020benchmarking} built a benchmark framework based on PyTorch and DGL\footnote{DEEP GRAPH LIBRARY: \href{https://www.dgl.ai}{https://www.dgl.ai}} 
that evaluates models on graph classification and graph regression tasks with an unified model evaluation protocol. 
They do apply training tricks, such as batch normalization, residual connections, and graph size normalization, to GL-GNNs to measure their effects. 
But all models being evaluated with the protocol are subject to the same training regime.
Similarly, in addition to the large-scale graph datasets, OGB \cite{hu2020open} provides a standard model evaluation protocol that includes a unified way to load and split data, the model evaluation itself, plus the cross-validations. Recently, Zhu \textit{et al.} \cite{zhu2021an} provides a benchmark framework for graph contrastive learning.

%% file: 9-Applications.tex
\section{Downstream tasks and Applications}

This section introduces the mainstream downstream tasks of graph-level learning and their corresponding applications. 

\noindent{\textbf{Graph Generation.}} This task aims to generate new graphs that have specific proprieties based on a series of graphs. 
Graph generation has a broad application in the field of biochemistry. For instance, \emph{drug development} involves experimenting with a tremendous number of molecule arrangements, but, through graph generation, the overall time and investment required to do this work can be reduced \cite{vamathevan2019applications}. Similarly, \emph{molecule generation} \cite{Shi*2020GraphAF:,morgan1965generation} has been used to explore new catalysts \cite{chanussot2021open}. 
Sanchez \textit{et al.} \cite{sanchez2020learning} applied graph generation into \emph{physical systems modeling} to simulate real-world particle motions. \emph{Scene graph generation} \cite{xu2017scene,ren2015faster} can be used to understand the scene of images and generate abstraction for images to summarize the relationship among objects in an image. Most recently, a few works \cite{allamanis2018learning,dinella2020hoppity} have employed graph generation for \emph{program debugging}, which modifies the nodes (i.e., variables or functions) and links in the program flow graph to fix bugs.

\noindent \textbf{Graph Classification.} 
The goal of graph classification is to learn the mapping relationship between graphs and corresponding class labels and predict the labels of unseen graphs. Graph classification is a critical graph-level learning task with a range of applications. 
For example, classifying molecular graphs \cite{wang2022molecular,fang2022geometry} can be used to determine anti-cancer activity, toxicity, or the mutagenicity of molecules. Classifying protein graphs \cite{borgwardt2005protein} can help to identify proteins with specific functions, such as enzymes. By converting texts to graphs in which nodes denote words and edges are the relationships between words, \emph{text categorization} \cite{Rousseau2015textgraphclass,peng2018large} can distinguish documents with different topics. By the same token, pixels in images can be regarded as nodes and adjacent pixels are linked to yield graphs for \emph{image recognition} \cite{chen2019multi,wu2015multi}. This task can be extended to \emph{medical diagnosis} to deal with computed tomography scans \cite{hao2022uncertainty} and clinical images \cite{wu2020learning}. In addition, graph classification can also be used for \emph{online product recommendation} \cite{wu2014bag} and \emph{fake news detection} \cite{dou2021user,silva2021embracing}. 
Recently, it has been impressive to see that graph-level learning can deliver \emph{IQ tests} \cite{Wang2020Abstract} that select graphs with a specific style from a group of graphs based on the style learning on the other group of graphs.

\noindent \textbf{Graph Comparison.} 
This task involves measuring the distance or similarity of pair-wise graphs in a graph dataset. The applications of this task include: \emph{semantic inference}, which infers text-document affiliations \cite{haghighi2005robust}; matching images with texts describing the same thing \cite{liu2020graph}; \emph{semantic metrics}, which measures the semantic similarity between texts \cite{ramage2009random,hughes2007lexical}; and \emph{cross-language information retrieval}, which seeks information in a language context that is different from the query \cite{xu-etal-2019-cross-lingual,monz2005iterative}.

\noindent \textbf{Graph Regression.} This task aims to predict the continuous proprieties of graphs. Taking molecules as examples, graph regression can predict different molecular proprieties related to the tightness of chemical bounds, fundamental vibrations, the state of electrons in molecules, the spatial distribution of electrons in molecules, and so on \cite{gilmer2017neural, jiang2021could, stokes2020deep}. Hence, the most promising application of graph regression is \emph{drug discovery}. In addition, employing graph regression to predict ratings or avenues of films is feasible.

\noindent\textbf{Subgraph Discovery.} 
This is the task of detecting discriminative substructures in a graph dataset. Subgraph Discovery can be applied to \emph{molecular structure search} \cite{ralaivola2005graph,duvenaud2015convolutional}, which explores the functional structures in chemical compounds, or to \emph{social event detection} \cite{shao2017efficient}, where subgraph discovery can be used to detect the substructures that represent great events in a series of social networks.

\noindent \textbf{Applying Complex Scenarios.} In addition to simple downstream tasks, researchers have extended graph-level learning to some complex scenarios. 
For instance, \emph{multi-view GL} targets learning in scenarios where an object is described by multiple graphs (i.e., multi-graph-views).
In multi-view GL, practitioners mine information from each single-graph-view and then strategically fuse information from all graph-views \cite{wu2016multi,wu2017multiple}.
\emph{Multi-task GL} \cite{pan2015joint,pan2016task} is generally used to optimize multiple related tasks; hence, it focuses on detecting the discriminative features across all the different tasks. 
In real-world scenarios, there are a vast number of unlabeled graphs that go unused since most GL techniques require learning from labeled information.
Consequently, \emph{semi-supervised GL} \cite{kong2010semi} was developed, which can learn from a dataset containing only a few labeled graphs and very many unlabeled graphs. Likewise, \emph{positive and unlabeled GL} \cite{wu2014PU,wu2016positive} only requires a few labeled graphs in one class along with other unlabeled graphs.
In terms of dynamic scenarios, there are also applications that record changing graphs over time as graph streams.
For example, a paper and its references can be regarded as a citation graph, and a graph stream can be produced of citations in chronological order of the corresponding papers.
\emph{Graph stream GL} \cite{pan2014graphstream,aggarwal2011classification}, for example, is specifically designed for graph stream data and mines valuable patterns from dynamic graph records.

%% file: 10-Future_Directions.tex
\section{Future Directions}
Although graph-level learning has gone through a long journey, there are still open issues that have been less explored.
In this section, we spotlight 12 future directions involving technical challenges and application issues of graph-level learning for readers to refer to.

\subsection{Neural Architecture Search (NAS) for GL-GNNs}

Existing GL-GNNs often have a complex architecture, consisting of a number of different components, e.g., multiple graph convolutions and graph pooling layers. 
GL-GNNs require careful parameter tuning to achieve optimal performance since most of them are non-convex. 
Hence, it is expensive to search for a well-performing architecture from among the bulk of optional components and their numerous parameters.

\emph{Opportunities:} Developing effective NAS methods to free researchers from the task of repeatedly searching for good architectures manually and, in turn, tuning the parameters is an urgent goal. 
By minimizing the entropy, Yang \textit{et al.} \cite{yang2023minimum} raised a dimension estimator, which can empower the GL-GNNs to automatically encode graphs into suitable dimensional embeddings.
Moreover, Knyazev \textit{et al.} \cite{knyazev2021parameter} modeled the search for an architecture as a graph in which each node represents a neural network layer or operation (e.g., a convolution layer) and each edge represents the connectivity between a pair of operations. Subsequently, GNNs can work on the constructed graphs to seek the optimal architecture. 
We argue that constructing an optimization goal based on knowledge of deep learning might be a practical way of providing an automatic NAS for various GL-GNNs.

\subsection{Geometrically Equivariant GL-GNNs}
In geometric graphs \cite{bronstein2021geometric}, each node is described by two vectors, i.e.,  a feature vector and a geometric vector. For example, in 3D molecule graphs, atoms are assigned geometric information such as speeds, coordinates, and spins which together comprise the geometric vector. 
Constructing GL-GNNs that can learn geometric graphs is an important part of modeling in chemistry and physics.

\emph{Opportunities:} GL-GNNs that can predict a set of geometric graphs need to be equivariant. For example, when inputting a geometric graph with a specific rotation into a GL-GNN, the corresponding output should reflect the same rotation. There are some algorithms about geometrically equivariant GL-GNNs. For example, Satorras \textit{et al.}'s \cite{satorras2021n} Equivariant Graph Neural Networks (EGNN) expands MPNNs aggregating both feature vectors and geometric vectors, while GemNet \cite{gasteiger2021gemnet} infuses more geometric information into the message passing mechanism, like dihedral angles. 
Both of these methods achieve state-of-the-art performance with 3D molecule prediction tasks.
For more details on this topic, we refer readers to \cite{han2022geometrically}.

\subsection{Self-explainable GL-GNNs}
Most algorithms for explaining the predictions of GL-GNNs are post-hoc (e.g., PGExplainer \cite{luo2020parameterized}), where the aim is to train a model to interpret a pre-trained GL-GNN. 
In other words, the training and explaining processes in GL-GNNs are independent.

\emph{Opportunities:} Miao \textit{et al.} \cite{miao2022interpretable} proposed that the separate prediction and explanation processes will inevitably lead to sub-optimal model performance. For example, the explanation model may detect substructures that have spurious correlations to the graph labels when interpreting predictions \cite{wu2022discovering}. Designing self-explaining GL-GNNs where the prediction and explanation components enhance each other should therefore be a fruitful future direction of research for the graph-level learning community.
\subsection{Informative Graph Pooling}

We categorized the existing pooling techniques into two families, i.e., global and hierarchical pooling (see Section \ref{Pooling}). 
The aim of the top-$k$ approaches \cite{zhang2018end,zhang2020structure}, which are among the most representative global pooling methods, is to select some nodes for the pooled graph. 
However, one cannot ensure that the redundancy of the selected nodes will be low. 
Further, the mechanism of the hierarchical family tends to smooth the node representations, which means the uninformative nodes tend to be selected for the pooled graphs \cite{mesquita2020rethinking}.

\emph{Opportunities:} Existing state-of-the-art graph pooling methods are not able to coarsen the original graph into a pooled graph with nodes of low redundancy. However, a pooled graph consisting of dissimilar nodes is critical for graph-level learning. For example, an atomic pair composed of different atoms can empower different proprieties to molecules. 
Traditional subgraph mining methods \cite{wu2014multi,wu2014bag} can then be used to identify the discriminative subgraphs of low redundancy as representative graphs. 
Hence, referring to the ideas of traditional subgraph mining for graph pooling methods that can identify informative nodes and/or subgraphs might yield feasible solutions to this problem.

\subsection{Graph-level Federated Learning}

Graph data are generally sourced from information collected by institutions. 
However, due to privacy considerations, graph data from different institutions is generally not used to jointly train graph-level learning models.
In practice, numerous graph-level learning techniques are data-hungry, especially the currently mainstream GL-GNNs.
Therefore, it is a practical topic to promote the joint training of graph-level learning models by different institutions using their respective graph data.

\emph{Opportunities:} Federated learning solves the data isolation problem, feeding data-driven machine learning models from different sources with rich amounts of data while maintaining privacy.
For example, Xie \textit{et al.} \cite{xie2021federated} proposed a federated learning framework specifically for GL-GNNs, where different GL-GNNs are trained based on different graph sets and sharing weights are learned by the GL-GNNs.
Graph-level federated learning is an emerging topic with great challenges.
In fact, a benchmark for this task has recently been released \cite{he2021fedgraphnn}.

\subsection{Graph-level Imbalance Learning} 
A machine learning model trained on the data with an imbalanced label distribution might be biased towards the majority classes. That is, with many samples and the minority classes consisting only of a small number of samples, the model may be under-fit. 
Representative tasks that need imbalance learning and must distinguish between samples from the majority and minority classes include anomaly detection \cite{zhang2022dual} and long-tail event detection \cite{agarwal2012catching}. 

\emph{Opportunities}: 
Although imbalanced learning has been a long-standing issue in deep learning, graph-level imbalance learning, especially with deep models, is underexplored. 
Wang \textit{et al.} \cite{wang2021imbalanced} over-sampled graphs in the minority class to relieve imbalance distributions between the majority and minority classes.
They also appended a self-consistency between the original and the augmented graphs. 
Over-sampling the minority samples is a traditional solution to imbalanced learning. However, this approach has been criticized for some shortcomings, such as over-fitting and changing the original distribution of the dataset. 
Additionally, minority graphs generally contain special substructures that are different from those in the majority graphs. 
Strengthening the structural awareness of the current graph-level learning tools could be a feasible way of overcoming this problem. 

\subsection{Graph-level Learning on Complex Graphs}

In this survey, almost all the investigated graph-level learning methods are assumed to work on fundamental graphs (i.e., unweighted and undirected graphs and their nodes and edges are homogeneous). 
This is because fundamental graphs are easy to understand and easy for models to handle. However, realistic graphs are usually complex. For example, the edges between actors and movies have a different meaning to the edges between two movies in multi-relational graphs. Collaborators on a paper can be linked together by a hyperedge (i.e., hypergraphs), while authors, papers, and venues can all be nodes in a citation network, even though they are distinct taxonomic entities (i.e., heterogeneous graphs), etc.

\emph{Opportunities:} Compared to highly developed graph-level learning on fundamental graphs, mining complex graphs still requires further development. 
For instance, most GL-GNNs for heterogeneous graphs rely on manually-defined meta-paths (i.e., a sequence of relations between nodes or edges) that are based on domain knowledge. 
However, defining meta-paths is not only expensive, it will not capture comprehensive semantic relationships \cite{hussein2018meta,yang2021interpretable}. 
Lv \textit{et al.} \cite{lv2021we} also raised the issue that, empirically, some heterogeneous GL-GNNs do not perform as well as simple GL-GNNs. 
In addition, it is hard to fairly evaluate hypergraph GL-GNNs since the hypergraphs are acquired from a range of different sources and built by a range of different construction approaches.
In conclusion, there are numerous worthwhile directions to explore when it comes to graph-level learning with complex graphs, such as benchmarking evaluation \cite{lv2021we} and datasets.

\subsection{Graph-level Interaction Learning}
Almost all the literature on graph-level learning treats each graph in a dataset as an independent sample.
However, considering the interactions between graphs should lead to challenging and highly novel research. 
For example, learning the interactions between graphs might be used to predict the chemical reactions when two compounds meet or to explore the effect of taking two or more drugs at the same time.

\emph{Opportunities:} Although this topic has strong practical implications for graph-level learning applications in biochemistry, it is still understudied.
So far, only a few GL-GNNs have been designed to tackle this topic and its related tasks.
DSS-GNN \cite{bevilacqua2021equivariant}, for instance, predicts the interactions between subgraphs located in a single graph, while Graph of Graphs Neural Network (GoGNN) \cite{10.5555/3491440.3491623} predicts chemical-chemical and drug-drug interactions. 
These two tasks own the off-the-shelf datasets, DDI \cite{Zitnik2018}, CCI \cite{kuhn2007stitch}, and SE \cite{Zitnik2018}.

\subsection{Graph-level Anomaly Detection}
The aim of anomaly detection is to identify objects that significantly deviate from the majority of other objects. 
However, when it comes to graph-structured data, almost all graph anomaly detection research focuses on detecting anomalous nodes in a single graph \cite{ma2021comprehensive}.

\emph{Opportunities:} Graph-level anomaly detection that identifies anomalous graphs in a graph dataset is a research topic of great value application-wise. 
For example, such a method could help to detect proteins with special functions from a large number of common protein structures. 
Some pioneering studies \cite{ma2022deep, qiu2022raising, zhao2021using} combine state-of-the-art GL-GNNs with traditional anomaly detection methods (e.g., one-class classification \cite{ruff2018deep}) to detect anomalous graphs in a graph dataset. 
However, these graph convolution operations were not specifically designed to detect anomalous graphs. 
Most graph convolution works like a low-pass filter \cite{balcilar2020analyzing} that smooths the anomalous information in a graph \cite{tang2022rethinking}. Hence, more analysis of the reasons behind anomalous graphs is needed and specific graph convolutions need to be proposed that are purposefully designed to detect anomalous graph information manifested in graph structures and/or attributes.

\subsection{Out-of-Distribution Generalization}

Out-of-distribution (OOD) learning improves a model's generalization ability.
It applies to scenarios where the test data does not have the same distribution as the training data. 
OOD settings can have two types of distribution shift, concept shift and covariate shift.
Concept shift refers to situations where the conditional distribution between the inputs and outputs differs from the training data to the test data. 
Covariate shift means that the test data has some certain features not shown in the training data.

\emph{Opportunities:} Almost all the graph-level learning algorithms assume that the training and the test data will have the same distribution. However, this I.I.D. (independent, identically distributed) assumption may be violated in some scenarios. 
For example, molecules with the same function may contain some different scaffolds. 
When the test data have a scaffold that has never appeared in training data, graph-level learning methods models will not perform nearly as well. The graph-level learning community has recently noticed this issue and has embarked on related research in response. 
Gui \textit{et al.} \cite{gui2022good} proposed a graph OOD learning benchmark. Inspired by invariant learning, Wu \textit{et al.} \cite{wu2022discovering} identified the casual subgraphs that are invariant across different distributions to improve the OOD generalization ability of GL-GNNs. Similarly, Bevilacqua \textit{et al.} \cite{bevilacqua2021size} employed an inference model to capture approximately invariant causal graphs to improve the extrapolation abilities of GL-GNNs. 
In addition to invariant learning, many techniques such as meta-learning, data augmentation, and disentanglement learning are feasible for OOD learning. Combining these techniques with GL-GNNs is likely to be the future of achieving graph-level learning models with a strong OOD generalization capacity.

\subsection{Brain Graphs Analytics}

Brain networks, also known as connectomes, are maps of the brain where the nodes denote the brain regions of interest (ROIs) in the brain and the edges denote the neural connections between these ROIs. 
An important application of machine learning models pertaining to brain networks is to distinguish brains with neurological disorders from normal individuals and identify those regions of the brain that are the cause of brain disease.

\emph{Opportunities:} Existing graph-level learning algorithms especially GL-DNNs and GL-GNNs, tend to be over-parameterized for learning brain networks, which are usually sparse. Further, obtaining a brain network usually comes at a high cost, because it involves scanning an individual's brain and converting the neuro-image into a brain network. 
In addition, existing GL-DNNs and GL-GNNs cannot handle the correspondence of nodes between different graphs. However, different brain networks have the same ROIs, and node identities and ROIs are one-to-one correspondence \cite{sporns2022graph}. 
In summary, graph-level learning with brain networks requires models that are lightweight and can identify corresponding nodes between different graphs.

\subsection{Multi-Graph-Level Learning} Standard graph-level learning views each graph as an instance, which can be restrictive in practical applications. 
Considering a product that has multiple reviews on an online shopping page.
Each review can be represented as a graph of the textual semantics among the words.
To predict any properties of that online product, one needs to learn from review-based multi-graphs \textemdash that is, multi-graph-level learning.

\emph{Opportunities:} To the best of our knowledge, the current multi-graph-level learning algorithms are all traditional.
For example, Boosting based Multi-graph Graph Classification (bMGC) \cite{wu2014boosting} and Multi-Instance Learning Discriminative Mapping (MILDM) \cite{wu2018multigraph} are both subgraph mining methods that classify multi-graph objects by extracting informative subgraphs.
However, both two methods cannot use label information to guide the feature selection process. 
Developing deep learning models can better extract features for multi-graph-level learning via the label information.


%% file: 11-Conclusions.tex
\section{Conclusions}

This survey paper provides a comprehensive review of graph-level learning methods.
Due to the irregular structure of graphs, graph-level learning has long been a non-trivial task with related research spanning the traditional to the deep learning era. 
However, the community is eager for a comprehensive taxonomy of this complex field. 
In this paper, we framed the representative graph-level learning methods into four categories based on different technical directions. 
In each category, we provided a detailed discussion on, and comparison of, the representative methods.
We also discussed open-source materials to support research in this field, including datasets, algorithm implementations, and benchmarks, along with the most graph-level learning tasks and their potential industrial applications. 
Lastly, we raised 12 future directions based on currently open issues that would make valuable contributions to the graph-level learning community.


%% file: 12-appendix.tex
\section{Taxonomy of Graph-level Learning Techniques}\label{ap_tax}

\begin{figure*}[htbp!]
    \centering
    \includegraphics[width=0.9\textwidth]{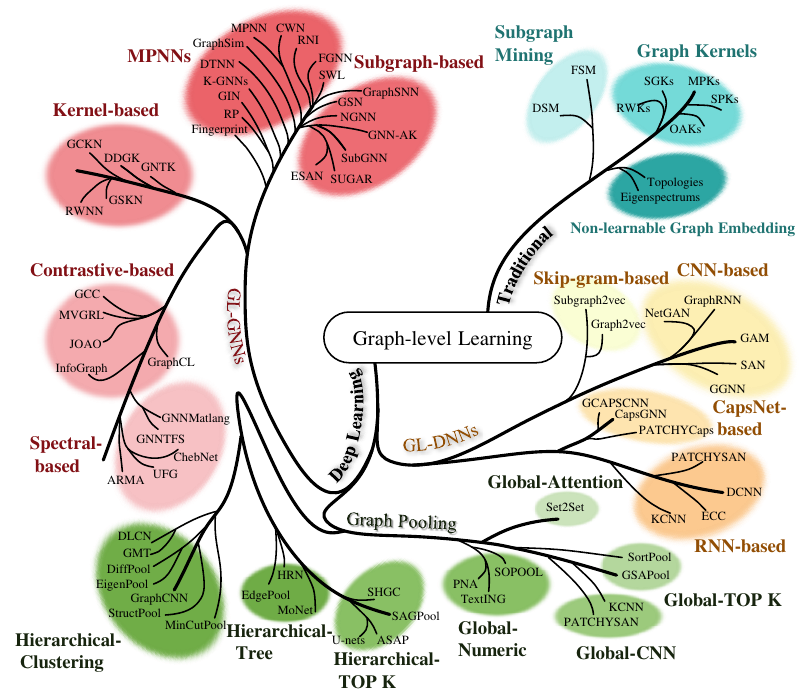}
    \caption{The taxonomy tree of graph-level learning techniques.}
    \label{tree}
\end{figure*}

The taxonomy tree in Fig. \ref{tree} depicts these four branches of graph-level learning with selected algorithms highlighted.

\section{Traditional Learning}\label{ap_tr}

All traditional graph-level learning publications discussed in this section are summarized in Table \ref{table_traditional}. 

\begin{table*}[]
\caption{Summary of Traditional Graph-level Learning Methods.}\label{table_traditional}
\scalebox{0.67}{ 
\begin{tabular}{cclllll}
\hline
Subsection                                                                  & Model                                                                                           & Year & Method                                            & Venue     & Language & Code Repository                                                                                                                 \\ \hline
\multirow{17}{*}{\begin{tabular}[c]{@{}c@{}}Graph \\ Kernels\end{tabular}}  & \multirow{5}{*}{\begin{tabular}[c]{@{}c@{}}Message\\ Passing\\ Kernels\end{tabular}}            & 2009 & NHK\cite{hido2009linear}                 & ICDM      & Python   & https://github.com/ysig/GraKeL                                                                                                  \\ \cline{3-7} 
                                                                            &                                                                                                 & 2011 & WL\cite{shervashidze2011weisfeiler}      & JMLR      & C++      & https://github.com/BorgwardtLab/graph-kernels                                                                                   \\ \cline{3-7} 
                                                                            &                                                                                                 & 2016 & PK\cite{neumann2016propagation}          & ML        & MATLAB   & https://github.com/marionmari/propagation\_kernels                                                                              \\ \cline{3-7} 
                                                                            &                                                                                                 & 2017 & Global-WL\cite{morris2017glocalized}     & ICDM      & C++      & https://github.com/chrsmrrs/glocalwl                                                                                            \\ \cline{3-7} 
                                                                            &                                                                                                 & 2019 & P-WL\cite{rieck2019persistent}           & ICML      & Python   & https://github.com/BorgwardtLab/P-WL                                                                                            \\ \cline{2-7} 
                                                                            & \multirow{2}{*}{\begin{tabular}[c]{@{}c@{}}ShortestPath\\ Kernels\end{tabular}}                 & 2005 & SPK\cite{borgwardt2005shortest}          & ICDM      & Python   & https://github.com/ysig/GraKeL                                                                                                  \\ \cline{3-7} 
                                                                            &                                                                                                 & 2017 & SPK-DS\cite{nikolentzos2017shortest}     & EMNLP     & -        & -                                                                                                                               \\ \cline{2-7} 
                                                                            & \multirow{3}{*}{\begin{tabular}[c]{@{}c@{}}Random\\ Walk\\ Kernels\end{tabular}}                & 2003 & RWK\cite{gartner2003graph}               & LNAI      & Python   & https://github.com/jajupmochi/graphkit-learn                                                                                    \\ \cline{3-7} 
                                                                            &                                                                                                 & 2004 & ERWK\cite{mahe2004extensions}            & ICML      & Python   & https://github.com/jajupmochi/graphkit-learn                                                                                    \\ \cline{3-7} 
                                                                            &                                                                                                 & 2010 & SOMRWK\cite{vishwanathan2010graph}       & JMLR      & Python   & https://github.com/ysig/GraKeL                                                                                                  \\ \cline{2-7} 
                                                                            & \multirow{4}{*}{\begin{tabular}[c]{@{}c@{}}Optimal\\ Assignment\\ Kernels\end{tabular}}         & 2005 & OAK\cite{frohlich2005optimal}            & ICML      & -        & -                                                                                                                               \\ \cline{3-7} 
                                                                            &                                                                                                 & 2013 & PS-OAK\cite{pachauri2013solving}         & NeurIPS   & Python   & https://github.com/zju-3dv/multiway                                                                                             \\ \cline{3-7} 
                                                                            &                                                                                                 & 2015 & GE-OAK\cite{johansson2015learning}       & KDD       & -        & -                                                                                                                               \\ \cline{3-7} 
                                                                            &                                                                                                 & 2015 & TAK\cite{schiavinato2015transitive}      & SIMBAD    & -        & -                                                                                                                               \\ \cline{2-7} 
                                                                            & \multirow{3}{*}{\begin{tabular}[c]{@{}c@{}}Subgraph\\ Kernels\end{tabular}}                     & 2009 & Graphlet\cite{shervashidze2009efficient} & AISTATS   & Python   & https://github.com/ysig/GraKeL                                                                                                  \\ \cline{3-7} 
                                                                            &                                                                                                 & 2010 & NSPDK\cite{costa2010fast}                & ICML      & Python   & https://github.com/fabriziocosta/EDeN                                                                                           \\ \cline{3-7} 
                                                                            &                                                                                                 & 2012 & SMK\cite{10.5555/3042573.3042614}        & ICML      & C++      & https://github.com/fapaul/GraphKernelBenchmark                                                                                  \\ \hline
\multirow{9}{*}{\begin{tabular}[c]{@{}c@{}}Subgraph \\ Mining\end{tabular}} & \multirow{3}{*}{\begin{tabular}[c]{@{}c@{}}Frequent\\ Subgraph\\ Mining\end{tabular}}           & 2000 & AGM\cite{inokuchi2000apriori}            & ECML PKDD & C++      & https://github.com/Aditi-Singla/Data-Mining                                                                                     \\ \cline{3-7} 
                                                                            &                                                                                                 & 2001 & FSG\cite{kuramochi2001frequent}          & ICDM      & C++      & https://github.com/NikhilGupta1997/Data-Mining-Algorithms                                                                       \\ \cline{3-7} 
                                                                            &                                                                                                 & 2002 & gSpan\cite{yan2002gspan}                 & ICDM      & Python   & https://github.com/betterenvi/gSpan                                                                                             \\ \cline{2-7} 
                                                                            & \multirow{6}{*}{\begin{tabular}[c]{@{}c@{}}Discrimina\\ -tive\\ Subgraph\\ Mining\end{tabular}} & 2008 & LEAP\cite{yan2008mining}                 & SIGMOD    & -        & -                                                                                                                               \\ \cline{3-7} 
                                                                            &                                                                                                 & 2009 & CORK\cite{thoma2009near}                 & SDM       & -        & -                                                                                                                               \\ \cline{3-7} 
                                                                            &                                                                                                 & 2010 & gMLC\cite{kong2010multi}                 & ICDM      & -        & -                                                                                                                               \\ \cline{3-7} 
                                                                            &                                                                                                 & 2010 & gSSC\cite{kong2010semi}                  & KDD       & -        & -                                                                                                                               \\ \cline{3-7} 
                                                                            &                                                                                                 & 2011 & gPU\cite{zhao2011positive}               & ICDM      & -        & -                                                                                                                               \\ \cline{3-7} 
                                                                            &                                                                                                 & 2014 & gCGVFL\cite{wu2014multi}                 & ICDM      & -        & -                                                                                                                               \\ \hline
\multicolumn{2}{c}{\multirow{6}{*}{\begin{tabular}[c]{@{}c@{}}Non-Learnable\\ Graph \\ Embedding\end{tabular}}}                                                               & 2017 & FGSD\cite{verma2017hunt}                 & NeurIPS   & Python   & https://github.com/vermaMachineLearning/FGSD                                                                                    \\ \cline{3-7} 
\multicolumn{2}{c}{}                                                                                                                                                          & 2018 & AWE\cite{ivanov2018anonymous}            & ICML      & Python   & https://github.com/nd7141/AWE                                                                                                   \\ \cline{3-7} 
\multicolumn{2}{c}{}                                                                                                                                                          & 2019 & LDP\cite{cai2018simple}                  & ICLR RLGM & Python   & https://github.com/Chen-Cai-OSU/LDP                                                                                             \\ \cline{3-7} 
\multicolumn{2}{c}{}                                                                                                                                                          & 2020 & SLAQ\cite{tsitsulin2020just}             & WWW       & Python   & \begin{tabular}[c]{@{}l@{}}https://github.com/google-research/google-research/tree/\\ master/graph\_embedding/slaq\end{tabular} \\ \cline{3-7} 
\multicolumn{2}{c}{}                                                                                                                                                          & 2021 & VNGE\cite{liu2021bridging}               & WWW       & Python   & https://github.com/xuecheng27/WWW21-Structural-Information                                                                      \\ \cline{3-7} 
\multicolumn{2}{c}{}                                                                                                                                                          & 2021 & A-DOGE\cite{sawlani2021fast}             & ICDM      & Python   & https://github.com/sawlani/A-DOGE                                                                                               \\ \hline
\end{tabular}
}
\end{table*}

\section{Graph-Level Deep Neural Networks (GL-DNNs)}\label{ap_dnn}

The representative GL-DNNs mentioned in this section are summarized in Table \ref{table_dnn}.

\begin{table*}[]
\centering
\caption{Summary of Graph-Level Deep Neural Networks (GL-DNNs).}\label{table_dnn}
\scalebox{0.65}{ 
\begin{tabular}{clllll}
\hline
Model                                                                                          & Year & Method                                                & Venue   & Language          & Code Repository                                                                                   \\ \hline
\multicolumn{1}{l}{\multirow{3}{*}{\begin{tabular}[c]{@{}l@{}}Skipgram\\ -Based\end{tabular}}} & 2016 & Subgraph2vec\cite{narayanan2016subgraph2vec} & KDD MLG & Python            & https://github.com/MLDroid/subgraph2vec\_tf                                                       \\ \cline{2-6} 
\multicolumn{1}{l}{}                                                                           & 2017 & Graph2vec\cite{narayanan2017graph2vec}       & KDD MLG & Python            & https://github.com/MLDroid/graph2vec\_tf                                                          \\ \cline{2-6} 
\multicolumn{1}{l}{}                                                                           & 2018 & GE-FSG\cite{nguyen2018learning}              & SDM     & Python            & https://github.com/nphdang/GE-FSG                                                                 \\ \hline
\multirow{5}{*}{\begin{tabular}[c]{@{}c@{}}RNN-\\ Based\end{tabular}}                          & 2016 & GGNN\cite{li2016gated}                       & ICLR    & Python-Tensorflow & https://github.com/Microsoft/gated-graph-neural-network-samples                                   \\ \cline{2-6} 
                                                                                               & 2018 & GAM\cite{lee2018graph}                       & KDD     & Python-Pytorch    & https://github.com/benedekrozemberczki/GAM                                                        \\ \cline{2-6} 
                                                                                               & 2018 & SAN\cite{zhao2018substructure}               & AAAI    & -                 & -                                                                                                 \\ \cline{2-6} 
                                                                                               & 2018 & NetGAN\cite{bojchevski2018netgan}            & ICML    & Python-Tensorflow & https://github.com/danielzuegner/netgan                                                           \\ \cline{2-6} 
                                                                                               & 2018 & GraphRNN\cite{you2018graphrnn}               & ICML    & Python-Pytorch    & https://github.com/snap-stanford/GraphRNN                                                         \\ \hline
\multirow{4}{*}{\begin{tabular}[c]{@{}c@{}}CNN-\\ Based\end{tabular}}                          & 2016 & PATCHYSAN\cite{niepert2016learning}          & ICML    & Python            & https://github.com/tvayer/PSCN                                                                    \\ \cline{2-6} 
                                                                                               & 2016 & DCNN\cite{NIPS2016_390e9825}             & NeurIPS & Python            & https://github.com/jcatw/dcnn                                                                     \\ \cline{2-6} 
                                                                                               & 2017 & ECC\cite{simonovsky2017dynamic}              & CVPR    & Python-Pytorch    & https://github.com/mys007/ecc                                                                     \\ \cline{2-6} 
                                                                                               & 2018 & KCNN\cite{nikolentzos2018kernel}             & ICANN   & Python-Pytorch    & \begin{tabular}[c]{@{}l@{}}https://github.com/giannisnik/\\ cnn-graph-classification\end{tabular} \\ \hline
\multirow{3}{*}{\begin{tabular}[c]{@{}c@{}}CapsNet-\\ Based\end{tabular}}                      & 2018 & GCAPSCNN\cite{verma2018graph}                & WCB     & Python            & https://github.com/vermaMachineLearning/Graph-Capsule-CNN-Networks                                \\ \cline{2-6} 
                                                                                               & 2019 & CapsGNN\cite{xinyi2018capsule}               & ICLR    & Python-Pytorch    & https://github.com/benedekrozemberczki/CapsGNN                                                    \\ \cline{2-6} 
                                                                                               & 2019 & PatchyCaps\cite{mallea2019capsule}           & Arxiv   & Python            & https://github.com/BraintreeLtd/PatchyCapsules                                                    \\ \hline
\end{tabular} }
\end{table*}

\subsection{CNN-based GL-DNNs}\label{ap_cnn}

\begin{figure*}[htbp!]
\begin{minipage}{1\linewidth}
\vspace{3pt}
\centerline{\includegraphics[width=\textwidth]{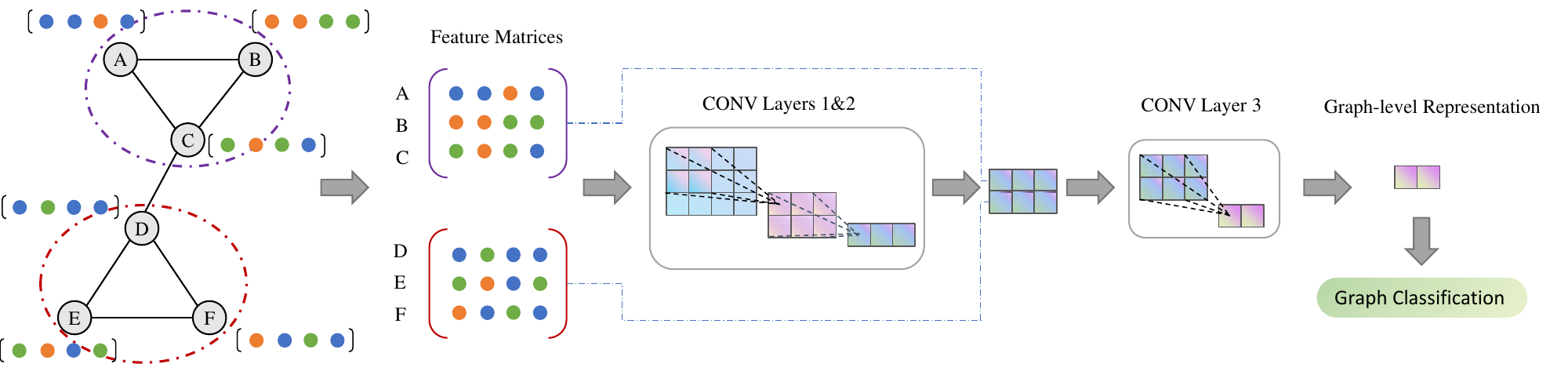}}
\centerline{(A)}
\end{minipage}
\begin{minipage}{1\linewidth}
\vspace{3pt}
\centerline{\includegraphics[width=\textwidth]{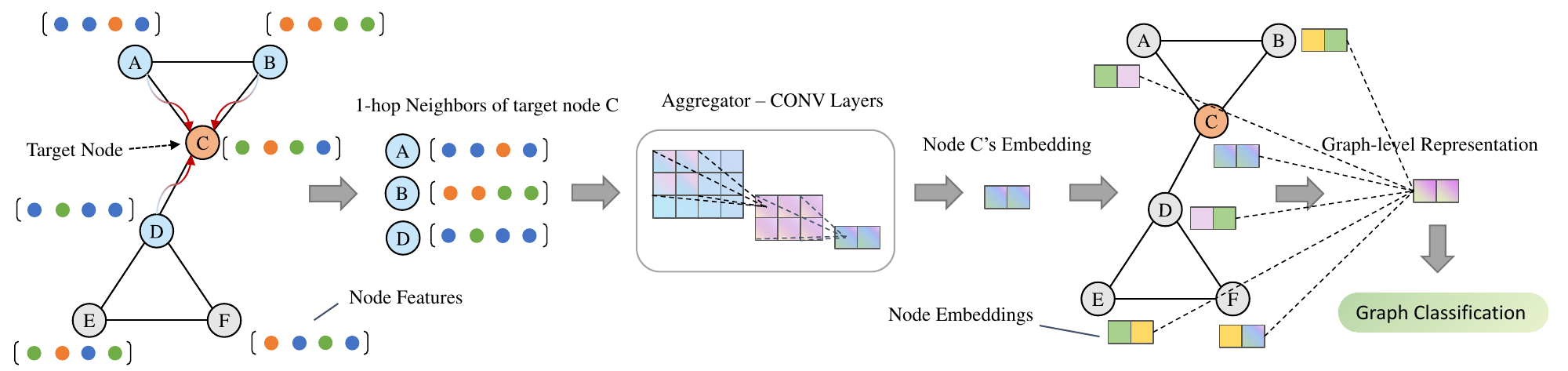}}
\centerline{(B)}
\end{minipage}
\caption{This figure shows two ways of tentative exploration by CNN-based GL-DNNs on graph-structured data.}    \label{fGLDNN}
\end{figure*}

There are two main branches of CNN-based graph-level learning. The first branch is illustrated in Fig. \ref{fGLDNN} (A), which sorts nodes and arranges the node features to form a concentration matrix as the grid-structured data for training the CNNs. 
As a second branch, researchers have developed a CNN-guided neural network version of an MPK, which is shown in Fig. \ref{fGLDNN} (B).

\section{GRAPH-LEVEL GRAPH NEURAL NETWORKS (GL-GNNS)}\label{ap_gnn}

\begin{table*}[htbp!]
\centering
\caption{Summary of Graph-Level Graph Neural Networks (GL-GNNs).}\label{table_gnn}
\scalebox{0.67}{ 
\begin{tabular}{clllll}
\hline
Model                                                                                           & Year & Method                                               & Venue   & Language          & Code Repository                                                                                                                 \\ \hline
\multirow{12}{*}{\begin{tabular}[c]{@{}c@{}}Message\\ Passing\\ Neural\\ Networks\end{tabular}} & 2015 & Fingerprint\cite{duvenaud2015convolutional} & NeurIPS & Python-Tensorflow & https://github.com/HIPS/neural-fingerprint                                                                                      \\ \cline{2-6} 
                                                                                                & 2016 & GraphSim\cite{battaglia2016interaction}     & NeurIPS & Python-Tensorflow & https://github.com/clvrai/Relation-Network-Tensorflow                                                                           \\ \cline{2-6} 
                                                                                                & 2017 & MPNN\cite{gilmer2017neural}                 & ICML    & Python-Pytorch    & https://github.com/priba/nmp\_qc                                                                                                \\ \cline{2-6} 
                                                                                                & 2017 & DTNN\cite{schutt2017quantum}                & NC      & Python-Tensorflow & https://github.com/atomistic-machine-learning/dtnn                                                                              \\ \cline{2-6} 
                                                                                                & 2019 & GIN\cite{xu2018powerful}                    & ICLR    & Python-Pytorch    & https://github.com/weihua916/powerful-gnns                                                                                      \\ \cline{2-6} 
                                                                                                & 2019 & K-GNNs\cite{morris2019weisfeiler}           & AAAI    & Python-Pytorch    & https://github.com/chrsmrrs/k-gnn                                                                                               \\ \cline{2-6} 
                                                                                                & 2019 & PPGN\cite{maron2019provably}                & NeurIPS & Python-Tensorflow & https://github.com/hadarser/ProvablyPowerfulGraphNetworks                                                                       \\ \cline{2-6} 
                                                                                                & 2019 & RP\cite{murphy2019relational}               & ICML    & Python-Pytorch    & https://github.com/PurdueMINDS/RelationalPooling                                                                                \\ \cline{2-6} 
                                                                                                & 2021 & FGNN\cite{azizian2021expressive}            & ICLR    & Python-Pytorch    & https://github.com/mlelarge/graph\_neural\_net                                                                                  \\ \cline{2-6} 
                                                                                                & 2021 & SWL\cite{bodnar2021weisfeiler_icml}       & ICML    & Python-Pytorch    & https://github.com/twitter-research/cwn                                                                                         \\ \cline{2-6} 
                                                                                                & 2021 & CWN\cite{bodnar2021weisfeiler_nips}        & NeurIPS & Python-Pytorch    & https://github.com/twitter-research/cwn                                                                                         \\ \cline{2-6} 
                                                                                                & 2021 & RNI\cite{DBLP:conf/ijcai/AbboudCGL21}       & IJCAL   & -                 & -                                                                                                                               \\ \hline
\multirow{7}{*}{\begin{tabular}[c]{@{}c@{}}Subgraph\\ -Based\end{tabular}}                      & 2020 & SubGNN\cite{AlsentzerFLZ20subgraph}         & NeurIPS & Python-Pytorch    & https://github.com/mims-harvard/SubGNN                                                                                          \\ \cline{2-6} 
                                                                                                & 2021 & SUGAR\cite{sun2021sugar}                    & WWW     & Python-Tensorflow & https://github.com/RingBDStack/SUGAR                                                                                            \\ \cline{2-6} 
                                                                                                & 2021 & NGNN\cite{zhang2021nested}                  & NeurIPS & Python-Pytorch    & https://github.com/muhanzhang/nestedgnn                                                                                         \\ \cline{2-6} 
                                                                                                & 2022 & GNN-AK\cite{zhao2022from}                   & ICLR    & Python-Pytorch    & https://github.com/LingxiaoShawn/GNNAsKernel                                                                                    \\ \cline{2-6} 
                                                                                                & 2022 & GraphSNN\cite{wijesinghe2021new}            & ICLR    & Python-Pytorch    & https://github.com/wokas36/GraphSNN                                                                                             \\ \cline{2-6} 
                                                                                                & 2022 & ESAN\cite{bevilacqua2021equivariant}        & ICLR    & Python-Pytorch    & https://github.com/beabevi/esan                                                                                                 \\ \cline{2-6} 
                                                                                                & 2022 & GSN\cite{bouritsas2022improving}            & TPAMI   & Python-Pytorch    & https://github.com/gbouritsas/GSN                                                                                               \\ \hline
\multirow{5}{*}{\begin{tabular}[c]{@{}c@{}}Kernel\\ -Based\end{tabular}}                        & 2019 & GNTK\cite{du2019graph}                      & NeurIPS & Python            & https://github.com/KangchengHou/gntk                                                                                            \\ \cline{2-6} 
                                                                                                & 2019 & DDGK\cite{al2019ddgk}                       & WWW     & Python-Tensorflow & \begin{tabular}[c]{@{}l@{}}https://github.com/google-research/google-research/\\ tree/master/graph\_embedding/ddgk\end{tabular} \\ \cline{2-6} 
                                                                                                & 2020 & GCKN\cite{chen2020convolutional}            & ICML    & Python-Pytorch    & https://github.com/claying/GCKN                                                                                                 \\ \cline{2-6} 
                                                                                                & 2020 & RWNN\cite{NEURIPS2020_ba95d78a}            & NeurIPS & Python-Pytorch    & https://github.com/giannisnik/rwgnn                                                                                             \\ \cline{2-6} 
                                                                                                & 2021 & GSKN\cite{long2021theoretically}            & WWW     & Python-Pytorch    & https://github.com/YimiAChack/GSKN                                                                                              \\ \hline
\multirow{5}{*}{\begin{tabular}[c]{@{}c@{}}Contrastive\\ -Based\end{tabular}}                   & 2020 & GraphCL\cite{NEURIPS2020_3fe23034}         & NeurIPS & Python-Pytorch    & https://github.com/Shen-Lab/GraphCL                                                                                             \\ \cline{2-6} 
                                                                                                & 2020 & InfoGraph\cite{sun2019infograph}            & ICLR    & Python-Pytorch    & https://github.com/fanyun-sun/InfoGraph                                                                                         \\ \cline{2-6} 
                                                                                                & 2020 & GCC\cite{qiu2020gcc}                        & KDD     & Python-Pytorch    & https://github.com/THUDM/GCC                                                                                                    \\ \cline{2-6} 
                                                                                                & 2020 & MVGRL\cite{hassani2020contrastive}          & ICML    & Python-Pytorch    & https://github.com/kavehhassani/mvgrl                                                                                           \\ \cline{2-6} 
                                                                                                & 2021 & JOAO\cite{you2021graph}                     & ICML    & Python-Pytorch    & https://github.com/Shen-Lab/GraphCL\_Automated                                                                                  \\ \hline
\multirow{5}{*}{\begin{tabular}[c]{@{}c@{}}Spectral\\ -Based\end{tabular}}                      & 2016 & ChebNet\cite{defferrard2016convolutional}   & NeurIPS & Python-Tensorflow & https://github.com/mdeff/cnn\_graph                                                                                             \\ \cline{2-6} 
                                                                                                & 2021 & GNNTFS\cite{levie2021transferability}       & JMLR    & -                 & -                                                                                                                               \\ \cline{2-6} 
                                                                                                & 2021 & GNNMatlang\cite{balcilar2021breaking}       & ICML    & Python-Tensorflow & https://github.com/balcilar/gnn-matlang                                                                                         \\ \cline{2-6} 
                                                                                                & 2021 & ARMA\cite{bianchi2021graph}                 & TPAMI   & Python-Pytorch    & https://github.com/dmlc/dgl/tree/master/examples/pytorch/arma                                                                   \\ \cline{2-6} 
                                                                                                & 2021 & UFG\cite{zheng2021framelets}                & ICML    & Python-Pytorch    & https://github.com/YuGuangWang/UFG                                                                                              \\ \hline
\end{tabular}
}
\end{table*}

We also illustrate the contrastive learning-based approaches (see Section \ref{contrastive GL-GNNs}). In addition, we investigate the expressivity (see Section \ref{expressivity}), generalizability (see Section \ref{Generalizability}), and explainability (see Section \ref{Explanation-GNN}) of GL-GNNs. Please refer to Table \ref{table_gnn} for the GL-GNNs discussed in this section.

\subsection{Contrastive Learning-Based GL-GNNs}\label{contrastive GL-GNNs}

Contrastive learning \cite{hjelm2018learning} is a data augmentation method that creates new, plausible instances by transposing existing data without affecting the semantics.
Investigating contrastive learning for GL-GNNs is significant since GL-GNNs are data-driven models that will always encounter bottlenecks given insufficiently labeled graphs. 

Graph Contrastive Learning (GraphCL) \cite{NEURIPS2020_3fe23034} defines four approaches to creating new instances as augmentation data: (1) node dropping, which randomly removes a proportion of nodes from the graph; (2) edge perturbation, which randomly adds or removes a certain percentage of edges from the graph; (3) feature masking, where some of the features of some nodes are randomly masked; and (4) subgraphs, where subgraphs are taken from the graph.
To be noticed, the newly-produced instances must be labeled as the same class as the source graph.
InfoGraph \cite{sun2019infograph}, for example, samples subgraphs $g_m$ from a source graph $\mathcal G$ as new instances.
A GL-GNN encoder $\mathcal H^{\phi}$ with some parameters $\phi$ is then used to generate graph-level representations of $g_m$ and $\mathcal G$, denoted as $\mathbf h^{\phi}_{g_m}$ and $\mathbf h^{\phi}_{\mathcal G}$.
InfoGraph's learning objective is to maximize the mutual information between $\mathbf h^{\phi}_{\mathcal G}$ and all $\mathbf h^{\phi}_{g_m}, g_m \in \mathcal G$.
This can be brought of as an evaluation of the statistical dependencies between two variables. Formally:
\begin{equation}
\mathcal H^{\phi}, \mathcal H^{\psi}=\underset{\phi, \psi}{\operatorname{argmax}} \sum_{\mathcal G \in \mathbb G} \frac{1}{|\{g_m\}|} \sum_{g_m \in \mathcal G} I_{\phi, \psi}\left(\mathbf h^{\phi}_{g_m} ; \mathbf h^{\phi}_{\mathcal G}\right),
\end{equation}
where $\mathcal H^{\psi}$ is the mutual information estimator with the parameters $\psi$, and $I_{\phi, \psi}\left(\cdot, \cdot\right)$ measures the mutual information.

Similarly, Graph Contrastive Coding (GCC) \cite{qiu2020gcc} samples subgraphs $g_1,...,g_M$ from the graph dataset $\mathbb G = \{\mathcal G_1,...,\mathcal G_N\}$ as new instances.
The embeddings of the subgraphs $g_m$ and graph $\mathcal G_n$ produced by the GL-GNN are denoted as $\mathbf h_{g_m}$ and $\mathbf h_{\mathcal G_n}$, respectively.
In the GL-GNN, a GCC employs InfoNCE loss \cite{oord2018representation} as the learning objective, that is:
\begin{equation}\label{infonce}
    \mathcal L = \sum_{\mathcal G_n \in \mathbb G} - \log \frac{\sum_{g_m \in \mathcal G_n} \exp\left(\mathbf h_{\mathcal G_n}^{\top} \mathbf h_{g_m}  / \tau \right)}
    {\sum_{i=0}^{M} \exp\left( \mathbf h_{\mathcal G_n}^{\top} \mathbf h_{g_i}  / \tau \right)},
\end{equation}
where $\tau$ is the temperature hyper-parameter. If $g_m \in \mathcal G_n$, the InfoNCE aims to maximize the similarity between $\mathbf h_{g_m}$ and $\mathbf h_{\mathcal G_n}$.
Otherwise, it separates $\mathbf h_{g_m}$ and $\mathbf h_{\mathcal G_n}$ as far away in the semantic space as possible.
Hassani \textit{et al.} \cite{hassani2020contrastive} extended graph contrastive learning to the multi-view scenario. Here, each graph view is regarded as an independent instance. This work maximizes the mutual information between a graph view and other views of the same graph. Recently, a contrastive-based GL-GNN MolCLR \cite{wang2022molecular} adopts three augmentation strategies (i.e., node drops, edge drops, and subgraph removal) for achieving benchmark results on 10 million unique molecules.

There are several ways to generate new instances for graph contrastive learning, which raises the question of how to choose the most suitable method for the dataset one is working with. 
Joint augmentation optimization (JOAO) \cite{you2021graph} was developed to address this challenge by automating the search for a proper graph data augmentation method. 
JOAO trains a probability matrix that can be iteratively updated to select the optimal data augmentation approach. Its performance is competitive.

\subsection{The expressivity of GL-GNNs}\label{expressivity}

As the cutting-edge technology for graph-level learning, people want to explore the power of GL-GNNs for distinguishing graphs \textemdash namely, they want to investigate the expressivity of GL-GNNs. 
Practitioners generally employ a representative MPK of the 1-WL algorithm \cite{weisfeiler1968reduction,shervashidze2011weisfeiler} to evaluate the expressivity of standard GL-GNNs (i.e., MPNNs) since MPNNs are the neural network versions of MPKs.
The intimate connection between GL-GNNs and 1-WL is exploited in the Graph Isomorphism Network (GIN) \cite{xu2018powerful}.
This framework shows the upper expressivity bound of an MPNN equals the 1-WL algorithm. 
Several research teams have subsequently proved that MPNNs equivalent to 1-WL can not distinguish some substructures in graphs (e.g., cycles, triangles, and Circulant Skip Links) \cite{arvind2020weisfeiler,vignac2020building,murphy2019relational}.
However, these indistinguishable substructures play a significant role in learning social network and chemical compounds graphs \cite{chen2020can}. 
To break the 1-WL expressivity limitation, the expressivity of GL-GNNs has been empowered through $K$-WL, convolution enhancement, and feature enrichment.

\subsubsection{$K$-WL} A complex variant of 1-WL is the $K$-WL algorithm, which identifies more substructures in graphs by relabeling a set of $K$ vertices. 
Morris \textit{et al.} \cite{morris2019weisfeiler} employed MPNNs dealing with $K$-dimensional tensors to apply $K$-WL by neural networks, that is $K$-GNN.
$K$-GNN achieved the expressivity approximately near but slightly weaker than the $K$-WL, but its computational cost increases exponentially with $K$ since it needs to calculate $K$-ranked tensors. 
To avoid processing high-dimensional tensors, Provably Powerful Graph Networks (PPGN) \cite{maron2019provably} adopts a variant of the 2-WL algorithm (i.e., 2-FWL \cite{cai1992optimal}) for designing GL-GNNs and achieves the expressivity over 3-WL.
Further, PPGN replaces the relabel function in 2-FWL with a matrix multiplication based on a single quadratic operation. 
Similarly, Folklore Graph Neural Networks (FGNN) \cite{azizian2021expressive} implements 2-FWL through matrix operations on tensors, pursuing the expressive power as 3-WL.
Despite these common efforts on $K$-WL equivalence GL-GNNs, the majority of them theoretically exceed 1-WL but do not empirically exceed 1-WL \cite{dwivedi2020benchmarking}. 
This weak performance by $K$-WL equivalent GL-GNNs is due to two main reasons which are explained next. 

\subsubsection{Convolution Enhancement} One reason for the failure of the $K$-WL approaches is that they break the local updates of MPNNs \cite{battaglia2018relational}, i.e., they no longer update vertices based on neighborhood information.
In practice, GL-GNNs require local updates to preserve the inductive bias property of the graph convolutions \cite{battaglia2018relational}.
Therefore, some researchers have explored more powerful GL-GNNs by upscaling the graph convolutions yet preserving the local updates.
Alon and Yahav \cite{alon2021on} noticed that the majority of GL-GNNs do not capture the long-range interactions between nodes because the number of convolutional layers is limited by over-smoothing issues \textemdash that is, the node embeddings tend to be similar after multiple aggregations. 
However, long-range interactions can influence the discriminativeness of graphs.
For example, methylnonane is identified by the atoms posited in the compound's two end sides. 
To address this issue, these researchers appended a fully linked adjacency matrix to the convolutional layer which aggregates the long-range information without violating any local updates.
Another powerful tool for enhancing the convolution layer is the matrix query language (MATLANG) \cite{brijder2019expressive,geerts2021expressive}. 
MATLANG strengthens a GL-GNN so that it can recognize more special substructures through its matrix operations, thereby reaching 3-WL expressivity.
Inspired by this work, Balcilar \textit{et al.} \cite{balcilar2021breaking} added MATLANG to the convolutional layer, while Greets and Reutter \cite{geerts2022expressiveness} evaluated the expressiveness of GL-GNNs through MATLANG instead of the 1-WL algorithm.

\subsubsection{Feature Enrichment} Another reason that $K$-WL methods outperform 1-WL in theory but do not achieve superior performance in experiments is that they ignore the role of node features.
As complementary information for graph structures, node features allow almost all graphs to be discriminated by 1-WL GL-GNNs.
Some practitioners have emphasized that considering node features can also improve the expressiveness of GL-GNNs, rather than just focusing on graph structures.
Murphy \textit{et al.} \cite{murphy2019relational} annotated a unique position descriptor for each node, that is, sorting all nodes. 
Adopting these position descriptors as node features can help a 1-WL GL-GNN to better handle featureless graphs and identify more structures.
To maintain permutation-invariant of graph-level learning, all permutations of node order should be enumerated and the average results should be taken.
Similarly, Colored Local Iterative Procedure (CLIP) \cite{10.5555/3491440.3491734} sorts the nodes in a substructure and gives them a local position descriptor for feature enrichment.
In addition, both Sato \textit{et al.} \cite{sato2020survey} and Abboud \textit{et al.} \cite{DBLP:conf/ijcai/AbboudCGL21} insert random features into nodes giving rise to stable and powerful GL-GNNs. 

\subsubsection{High-order Neural Networks} Recently, researchers have tried to improve the expressivity of GL-GNNs through algebraic topology.
This is because equipping graphs as a geometric structure can preserve more valuable properties. Cellular GL-GNNs \cite{bodnar2021weisfeiler_icml,bodnar2021weisfeiler_nips} perform MPNNs on cell complexes, an object including hierarchical structures (e.g., vertices, edges, triangles, tetrahedra). By replacing graphs with cell complexes, cellular GL-GNNs benefits from the better computational fabric for larger expressivity. Furthermore, sheaf neural networks \cite{bodnar2022neural,barbero2022sheaf} decorate a graph with a geometrical structure, sheaf, which constructs vector space for each node and edge and applies linear transformations among these spaces. 
A correct sheaf setting will allow an MPNN to pass messages along a richer structure.
Thus, linearly separate embeddings can be learned, which will enhance the expressive power of GL-GNNs.

\subsection{The generalizability of GL-GNNs}\label{Generalizability}

Real-world applications with graph data tend to involve complex scenarios, such as needing to train a model with only a small amount of labeled data that can ultimately perform well with a large-scale unlabeled test (i.e., size shift) or using only a few labeled training graphs to fit the bulk of unlabeled test graphs. The ability to generalize GL-GNNs is hence a crucial aspect of dealing with these challenges. 

\subsubsection{Size Generalization} 

Sinha \textit{et al.} \cite{sinha2020evaluating} stress the importance of generalizing GL-GNNs and presented evaluation criteria for this.
Xu \textit{et al.} \cite{xu2021how} theoretically explain that GL-GNNs have better size generalization capabilities than MLPs and can extrapolate trained models to test data that is different from the training set.
To this end, they presented a trick for MPNNs where the graph's vertices are updated by minimizing the aggregated information instead of through summation.
This trick improves generalization ability by altering the learning process from one that is non-linear to one that is linear.
Yehudai \textit{et al.} \cite{yehudai2021local} theoretically and empirically found the generalization ability of GL-GNNs as the discrepancies in substructures between large and small graphs grows. 
To solve this problem, they forced the GL-GNN to pay more attention to the substructures that are hidden in large unlabeled graphs but rarely appear in small labeled graphs.
SizeShiftReg \cite{buffelli2022sizeshiftreg} constrains GNNs to be robust to size-shift through a regularization approach.
SizeShiftReg coarsens the input graph and minimizes the discrepancy between the distribution of the original and coarsened graph embeddings.

\subsubsection{Few-shot Learning} In considering few-shot learning scenarios, Ma \textit{et al.} \cite{ma2020adaptive} found that there are also differences in the substructures between a few labeled graphs and a large number of unlabeled graphs.
This is because a statistical sample of the training data is too small to represent the substructural distributions of the whole dataset. 
Thus, they paid more attention to capturing substructures in unseen unlabeled graphs. 
Chauhan \textit{et al.} \cite{Chauhan2020FEW-SHOT} clusters graphs based on their spectral properties, to produce super-class graphs.
Graph-level representations can be learned from super-class graphs as they have excellent generalization.

\subsection{The explainability of GL-GNNs}\label{Explanation-GNN}

The black-box nature of deep neural networks limits the applicability of GL-GNNs to situations where trust is not an absolutely crucial requirement. 
Making GL-GNNs explain their predictions in a way that is more interpretable to humans is therefore of great significance to extending the research of GL-GNNs.
Studies on GL-GNNs need to shed insights into how they handle node features and topologies when it comes to predictions.
They also need to more clearly demonstrate how the models identify significant subgraphs and features.
Methods to explain GL-GNNs can be roughly divided into two categories.
One group involves methods that explain the prediction of each input graph;
the other group of methods captures common patterns in the predictions of a set of graphs as explanations.

\subsubsection{A Single Graph} There are three ways to understand GL-GNNs predictions based on a single graph: they can be perturbation-based, model-proxy-based, or gradient-based. 
Perturbation-based methods mask nodes, edges, or substructures in the input graph to generate new predictions.
These are then compared to the original input prediction to highlight the important features or structures influencing the GL-GNNs.

For example, GNNExplainer \cite{ying2019gnnexplainer} masks nodes and edges by changing the feature and adjacency matrices, to form masked graphs.
An input graph and its masked graphs are predicted by a trained GL-GNN, while GNNExplainer aims to find the masked graphs with maximized mutual information between its' prediction and the input's prediction.
This found masked graph is the one that preserves the most significant substructures to the GL-GNN's given prediction.
Alternatively, SubgraphX \cite{yuan2021explainability} samples a group of nodes' neighborhoods as subgraphs. 
A trained GL-GNN is then used to compute Shapley values \cite{kuhn1953contributions} for all the sampled subgraphs. These values represent each subgraph's contribution to the GL-GNN's prediction.
PGExplainer \cite{luo2020parameterized} trains an MLP to determine which edges are valuable to a GNN's prediction and then removes any irrelevant edges to form a new graph. Subsequently, the original and the newly-formed graph are fed into a trained GL-GNN so as to optimize an MLP by maximizing the mutual information between their predictions.

Model-proxy-based methods utilize a simpler surrogate model to approximate the predictions of GL-GNNs. 
PGM-Explainer \cite{vu2020pgm} adopts an explainable Bayesian network \cite{pearl1988markov} to calculate the relationship dependencies between nodes, so as to generate a probability graph that describes the input graph. 

Gradient-based approaches measure the importance of different input features by back-propagating the gradients of the neural networks.
Gradient-weighted Class Activation Mapping (Grad-CAM) \cite{pope2019explainability}, for example, takes the gradient value of each node embedding in a graph classification task as a measurement of the nodes' significance to the GL-GNN's prediction.
Grad-CAM then measures this subgraph's importance to the prediction by averaging the gradient values of all node embeddings from the subgraph.

\subsubsection{A Set of Graphs} What is common to all the above methods is that they can only learn independent explanations for each instance of a graph \cite{ying2019gnnexplainer,yuan2020explainability}.
However, often the predictions of GL-GNNs made by GL-GNNs are based on a set of graphs. 
Thus, understanding the rules or graph patterns that a GL-GNN mines from a set of graphs can provide high-level and generic insights into the explainability of GL-GNNs.
XGNN \cite{yuan2020xgnn} employs a reinforcement learning guided graph generator that generates a graph pattern for different graphs in the same class.
The graph generator is trained via policy gradient to maximize the certain label prediction \cite{sutton1999policy}. Recently, Azzolin \textit{et al.} \cite{azzolin2023global} set a kernel function between extracted subgraphs and trainable prototypes and feed the kernel value into an MLP for prediction.
Finally, the trainable prototypes are assumed as the global explanation of a set of graphs.

\section{Graph Pooling}\label{ap_Pooling}

\begin{table*}[]
\centering
\caption{Summary of Graph Pooling.}\label{table_graphpooling}
\scalebox{0.68}{ 
\begin{tabular}{clllll}
\hline
Model                                                                               & Year                     & Method                                        & Venue   & Language          & Code Repository                                                                                                                              \\ \hline
\multirow{3}{*}{\begin{tabular}[c]{@{}c@{}}Global-\\ Numeric\end{tabular}}          & 2020                     & PNA\cite{NEURIPS2020_99cad265}      & NeurIPS & Python-Pytorch    & https://github.com/lukecavabarrett/pna                                                                                                       \\ \cline{2-6} 
                                                                                    & 2020                     & TextING\cite{zhang2020every}         & ACL     & Python-Tensorflow & https://github.com/CRIPAC-DIG/TextING                                                                                                        \\ \cline{2-6} 
                                                                                    & 2020                     & SOPOOL\cite{wang2020second}          & TPAMI   & Python-Pytorch    & https://github.com/divelab/sopool                                                                                                            \\ \hline
\begin{tabular}[c]{@{}c@{}}Global-\\ Attention\end{tabular}                         & \multicolumn{1}{c}{2016} & Set2Set\cite{set2set}                & ICLR    & Python-Pytorch    & https://github.com/pyg-team/pytorch\_geometric                                                                                               \\ \hline
\multirow{2}{*}{\begin{tabular}[c]{@{}c@{}}Global-\\ CNN\end{tabular}}              & 2016                     & PATCHYSAN\cite{niepert2016learning}  & ICML    & Python            & https://github.com/tvayer/PSCN                                                                                                               \\ \cline{2-6} 
                                                                                    & 2018                     & KCNN\cite{nikolentzos2018kernel}     & ICANN   & Python-Pytorch    & \begin{tabular}[c]{@{}l@{}}https://github.com/giannisnik/\\ cnn-graph-classification\end{tabular}                                            \\ \hline
\multirow{2}{*}{\begin{tabular}[c]{@{}c@{}}Global-\\ Top K\end{tabular}}            & 2018                     & SortPool\cite{zhang2018end}          & AAAI    & Python-Pytorch    & https://github.com/muhanzhang/pytorch\_DGCNN                                                                                                 \\ \cline{2-6} 
                                                                                    & 2020                     & GSAPool\cite{zhang2020structure}     & WWW     & Python-Pytorch    & https://github.com/psp3dcg/gsapool                                                                                                           \\ \hline
\multirow{7}{*}{\begin{tabular}[c]{@{}c@{}}Hierarchical-\\ Clustering\end{tabular}} & 2014                     & DLCN\cite{ICLR2014BRUNA}             & ICLR    & -                 & -                                                                                                                                            \\ \cline{2-6} 
                                                                                    & 2015                     & GraphCNN\cite{henaff2015deep}        & Arxiv   & Python-Tensorflow & https://github.com/mdeff/cnn\_graph                                                                                                          \\ \cline{2-6} 
                                                                                    & 2018                     & DiffPool\cite{ying2018hierarchical}  & NeurIPS & Python-Pytorch    & https://github.com/RexYing/diffpool                                                                                                          \\ \cline{2-6} 
                                                                                    & 2019                     & EigenPool\cite{ma2019graph}          & KDD     & Python-Pytorch    & https://github.com/alge24/eigenpooling                                                                                                       \\ \cline{2-6} 
                                                                                    & 2020                     & StructPool\cite{yuan2020structpool}  & ICLR    & Python-Pytorch    & https://github.com/Nate1874/StructPool                                                                                                       \\ \cline{2-6} 
                                                                                    & 2020                     & MinCutPool\cite{bianchi2020spectral} & ICML    & Python-Tensorflow & \begin{tabular}[c]{@{}l@{}}https://github.com/FilippoMB/Spectral-Clustering\\ -with-Graph-Neural-Networks-for-Graph-Pooling\end{tabular}     \\ \cline{2-6} 
                                                                                    & 2021                     & GMT\cite{baek2020accurate}           & ICLR    & Python-Pytorch    & https://github.com/JinheonBaek/GMT                                                                                                           \\ \hline
\multirow{4}{*}{\begin{tabular}[c]{@{}c@{}}Hierarchical-\\ Top K\end{tabular}}      & 2018                     & SHGC\cite{cangea2018towards}         & Arxiv   & Python-Tensorflow & https://github.com/HeapHop30/hierarchical-pooling                                                                                            \\ \cline{2-6} 
                                                                                    & 2019                     & U-Nets\cite{gao2019graph}            & ICML    & Python-Pytorch    & https://github.com/HongyangGao/Graph-U-Nets                                                                                                  \\ \cline{2-6} 
                                                                                    & 2019                     & SAGPool\cite{lee2019self}            & ICML    & Python-Pytorch    & https://github.com/inyeoplee77/SAGPool                                                                                                       \\ \cline{2-6} 
                                                                                    & 2020                     & ASAP\cite{ranjan2020asap}            & AAAI    & Python-Pytorch    & https://github.com/malllabiisc/ASAP                                                                                                          \\ \hline
\multirow{4}{*}{\begin{tabular}[c]{@{}c@{}}Hierarchical-\\ Tree\end{tabular}}       & 2017                     & MoNet\cite{monti2017geometric}       & CVPR    & Python-Pytorch    & https://github.com/dmlc/dgl/tree/master/examples/mxnet/monet                                                                                 \\ \cline{2-6} 
                                                                                    & 2019                     & EdgePool\cite{diehl2019edge}         & Arxiv   & Python-Pytorch    & \begin{tabular}[c]{@{}l@{}}https://github.com/pyg-team/pytorch\_geometric\\ /blob/master/torch\_geometric/nn/pool/edge\_pool.py\end{tabular} \\ \cline{2-6} 
                                                                                    & 2022                     & HRN\cite{wu2021structural}           & IJCAL   & Python            & https://github.com/Wu-Junran/HierarchicalReporting                                                                                           \\ \cline{2-6} 
                                                                                    & 2022                     & SEP-G\cite{wu2022structural}         & ICML    & Python            & https://github.com/Wu-Junran/SEP                                                                                                             \\ \hline
\end{tabular}
}
\end{table*}

Table \ref{table_graphpooling} summarizes the graph pooling approaches introduced in this section.
Moreover, we introduce some recent investigations about the efficacy of this newly-emerging technique (see Section \ref{effectiveness}).

\subsection{The effectivity of Graph Pooling}\label{effectiveness}

As a downstream summarization component of GNNs, graph pooling has attracted a surge of research interest. 
However, since graph pooling is so new, much work is required to investigate the effectiveness of all the various graph pooling algorithms.
Mesquita \textit{et al.} \cite{mesquita2020rethinking} conducted controlled experiments to empirically evaluate the effectiveness of clustering-based hierarchical graph pooling. 
First, they adopted two opposite strategies for guiding some clustering-based hierarchical graph pooling processes \textemdash specifically, clustering each of non-adjacent and adjacent nodes.
The final results not only show that the two strategies are comparable, they also indicate that off-the-shelf clustering algorithms, which tend to cluster adjacent nodes, fail to improve graph pooling.
As part of the experiments, Mesquita and colleagues also replaced the learnable assignment matrix in DiffPool \cite{ying2018hierarchical} with an immutable probability assignment matrix: uniform, normal and Bernoulli distributions were selected. 
The experimental results verify that the performance of fixed-probability-assignment-matrix-guided graph pooling is not weaker than that of DiffPool.
Overall, they concluded that the current clustering-based hierarchical pooling may not be particularly effective and matched this will a call for more sanity checks and ablation studies of the current graph pooling algorithms to fairly evaluate their contributions. 
Another study on Pooled Architecture Search (PAS) \cite{wei2021pooling} was dedicated to investigating the effectiveness of graph pooling \textemdash this time with different datasets.
The results of the study show that the effectiveness of graph pooling algorithms is data-specific, that is to say, different input data needs to be handled by a suitable graph pooling algorithm.
For this reason, PAS includes a differentiable search method to select the most appropriate graph pooling algorithm for the given input data.

\section{Downstream tasks and Applications}

\begin{figure*}[htbp!]
    \centering
    \includegraphics[width=1\textwidth]{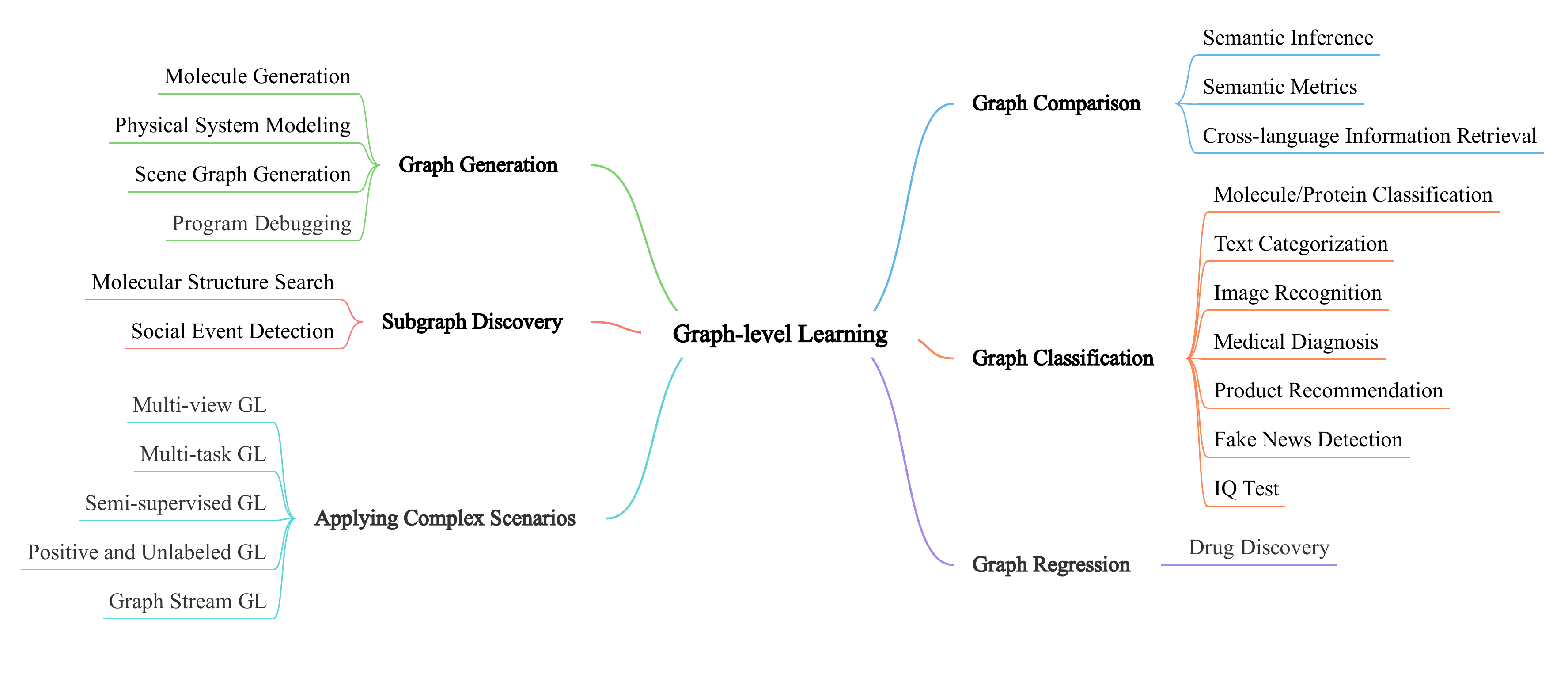}
    \caption{An overview of graph-level learning downstream tasks and their practical applications.}
    \label{fapplication}
\end{figure*}

Fig. \ref{fapplication} summarizes some common downstream tasks and applications in graph-level learning.

\section{Future Directions}

\subsection{Graph-level Fairness Learning}
The bias in data can easily lead to issues with fairness, where machine learning models make discriminatory predictions towards certain demographic groups based on sensitive attributes such as race. 
One feasible solution to debiasing the data is to conduct a competitive game between a biased and a debiased encoder.
The game is won when the fairness-aware debiaser is able to cheat its competitor \cite{bose2019compositional,masrour2020bursting}. 
Other algorithms add constraints to the loss function to counterbalance model performance with fairness \cite{kang2020inform,li2020dyadic}.

\emph{Opportunities:} 
Most work on improving the fairness of models have involved node-level tasks and single graphs \cite{dong2022edits}.
However, injecting an awareness of fairness into graph-level learning algorithms is also critical work. 
Some graph-level learning tasks, such as disease prediction and fraud detection, demand fair results if they are to accurately guide people's decision-making. 
One challenge to be overcome in attempting to make graph-level learning fair is that the representative GL-GNNs, i.e., MPNNs, will tend to produce unfair predictions in the face of data bias because the message passing mechanisms actually spread the bias via neighborhood structures \cite{dai2021say}. 
We refer readers who are interested in this topic to \cite{dai2022comprehensive}, which gives an exhaustive introduction to fairness learning with graph-structured data.

\bibliographystyle{ieeetr}
\bibliography{acm}